\documentclass{article}
\usepackage{microtype}
\usepackage{amsfonts}
\usepackage{amsmath}
\usepackage{graphicx}
\usepackage{multicol, multirow}   
\usepackage{subcaption}
\usepackage{mwe}
\usepackage{booktabs} 
\newbox{\bigpicturebox}

\usepackage{hyperref}
\usepackage{comment}
\usepackage[dvipsnames]{xcolor}

\usepackage[accepted]{icml2022}

\icmltitlerunning{Stochastic Volatility Models as Gaussian Processes}

\begin{document}

\twocolumn[
\icmltitle{Volatility Based Kernels and Moving Average Means \\
for Accurate Forecasting with Gaussian Processes}

\icmlsetsymbol{equal}{*}

\begin{icmlauthorlist}
\icmlauthor{Gregory Benton}{equal,nyu}
\icmlauthor{Wesley J. Maddox}{equal,nyu}
\icmlauthor{Andrew Gordon Wilson}{nyu}
\end{icmlauthorlist}

\icmlaffiliation{nyu}{New York University}

    \icmlcorrespondingauthor{Greg Benton}{gwb260@nyu.edu}

\icmlkeywords{Machine Learning, ICML}

\vskip 0.3in
]

\printAffiliationsAndNotice{\icmlEqualContribution} 

\begin{abstract}
A broad class of stochastic volatility models are defined by systems of stochastic differential equations. While these models have seen widespread success in domains such as finance and statistical climatology, they typically lack an ability to condition on historical data to produce a true posterior distribution.
To address this fundamental limitation, we show how to re-cast a class of stochastic volatility models as a hierarchical Gaussian process (GP) model with specialized covariance functions. This GP model retains the inductive biases of the stochastic volatility model while providing the posterior predictive distribution given by GP inference.
Within this framework, we take inspiration from well studied domains to introduce a new class of models, \emph{Volt} and \emph{Magpie}, that significantly outperform baselines in stock and wind speed forecasting, and naturally extend to the multitask setting.
\end{abstract}
\section{Introduction}
\label{sec: intro}

Gaussian processes (GP) have had significant success in time series modeling, making them strong candidates for the challenging tasks of modeling and forecasting time dependent financial and climatological data.
However, building GPs has historically relied on selecting out-of-the-box kernels and mean functions that make assumptions that do not hold for all cases; for example, Mat\'ern and RBF kernels assume stationarity of the data, and constant or linear means assume that the underlying trends in the data are not time-varying.
While the quest to build more flexible GPs has led to a significant amount of research into kernel functions such as spectral mixture and deep kernels, these kernels have been primarily developed for general problems, rather than constructing a kernel from prior knowledge about the task at hand. 
In this work, we approach GP modeling by building on domain knowledge to construct a novel set of kernel and mean functions with inductive biases well suited for forecasting in domains such as finance and climatology, where the data evolves stochastically through time.

\begin{figure}[t]
    \centering
    \includegraphics[width=0.95\linewidth]{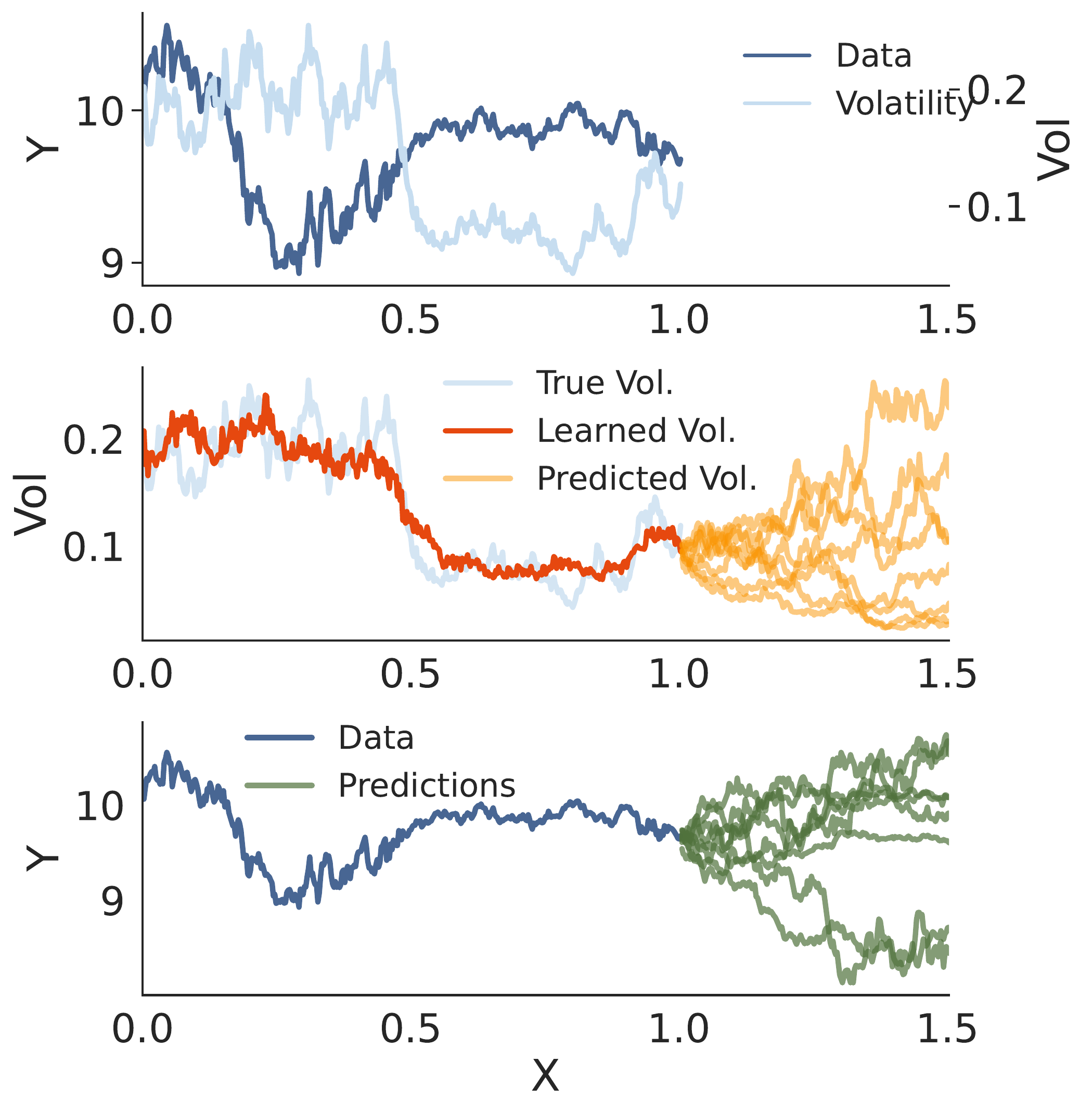}
    \caption{An overview of stochastic volatility, Volt, and forecasting. \textbf{Top:} the observed data over and the corresponding volatility path. \textbf{Middle:} the learned volatility from the data, and volatility forecasts. \textbf{Bottom:} the data over $1$ year and forecasts, with each sample path corresponding to a distinct sample from the volatility forecast.}
    \label{fig: intro}
\end{figure}

Both financial and climatological time series are nonstationary, and are characterized by having time-varying and stochastic \emph{volatilities}, or degrees of variation.
Motivated by stochastic volatility (SV) models, we develop a hierarchical GP model with specialized kernels, terming the model \emph{Volt} \citet{bollerslev1986generalized,wilson_copula_2010,hagan2002managing}.
Volt uses forecasts of volatility to specify the covariance structure over future data observations. By considering not only a single volatility forecast, but a distribution of volatility forecasts, we induce a distribution over covariance functions in the data domain. Accounting for uncertainty in volatility and propagating it to our data forecasts yields projected distributions that are well calibrated to the data, providing critical tools for understanding risk levels and simulating potential outcomes. For further information on stochastic volatility see Appendix \ref{app: tutorial}.

Figure \ref{fig: intro} provides a graphical representation of the hierarchical GP model described by Volt. Given a set of observations (top row) we infer a volatility path over those returns (middle row), and form a hierarchical GP model where the first GP models volatility, and the second GP is used to forecast distributions over the data given samples from the volatility GP (bottom row). 

The covariance structure described by Volt provides a faithful representation of the uncertainty in forecasts, but overlooks the mean function of the data space GP, which is a powerful tool for capturing trends in data. To that end, we jointly introduce Moving Average Gaussian Processes, or \emph{Magpie}, in which we replace the standard parametric mean function in GPs with a moving average. Moving averages are a widely used technique in domains such as climatology and finance \citep{nau2014forecasting}. By joining the trend fitting capabilities of moving averages with the probabilistic framework of GPs we can produce forecasts that are both accurate and have calibrated uncertainties.

While Volt and Magpie can be used separately, we present them as a single work because it is specifically their \emph{combination} that solves challenging forecasting problems.
In time evolving domains like stock prices or wind speeds, the inherent randomness of the processes prevents us from producing accurate point estimates far out into the future, and we need just accuracy, but uncertainty that is faithful to the stochasticity of the data.
For this reason one needs both the accurate trend capture provided by Magpie, and the accurate uncertainty representation provided by Volt 

Our key contributions are as follows:
\begin{itemize}
    \item Deriving a hierarchical GP model, Volt, inspired by stochastic volatility models that produces calibrated forecasts of stochastic time series (Section \ref{sec: methods}).
    \item Describing a simple but powerful mean function, Magpie, that enables Gaussian process models to accurately forecast trends (Section \ref{sec: methods}).
    \item Using Volt and Magpie to produce highly calibrated forecasts in financial and climatological domains (Section \ref{sec:forecasting}).
    \item Extending our procedure to multitask problems by accounting for correlations in both volatility and price across different financial assets and different spatial locations (Section \ref{sec: multitask}).
    \item We make code available \href{https://github.com/g-benton/Volt}{here}.
\end{itemize}

\section{Related Work}

Early autoregressive approaches to modeling the volatility of time series returns such as GARCH have seen widespread success \citep{bollerslev1986generalized}. These approaches typically view the volatility process as a time-evolving series, and are effective for inferring and forecasting volatility, but do not typically interface directly with a model over data as we have with Volt.  

Volatility models have been extended to use both neural networks or Gaussian processes as their base components. 
For example, \citet{cao2020neural} use a multi-layer perceptron to estimate volatility surfaces while \citet{luo2018neural} use RNNs with rollouts to forecast volatility into the future but only considered one-step lookahead price forecasts. 
\citet{wilson_copula_2010} use Gaussian processes to parameterize the volatility using Laplace approximations and MCMC sampling introducing the Gaussian process copula volatility model (GPCV), while \citet{wu2014gaussian} used GP state space models and particle filters to estimate volatility.
Similar to \citet{wilson_copula_2010}, \citet{lazaro2011variational} used Gaussian processes to parameterize volatility models with an exponential link, but used a highly structured variational approximation for inference.
\citet{liu2020overview} used multi-task Gaussian processes to forecast volatility into the future, applying their models to foreign exchange currency returns, again with one-step lookahead forecasts in price.
Crucially, predicting volatility alone does yield a straightforward path to forecast data, which is our central aim with Volt.
Furthermore, Volt builds off of the GPCV, but other volatility estimation methods such as the ones described here could also be used.

Stochastic volatility models such as the Heston model \citep{heston1993closed} and SABR \citep{hagan2002managing}, treat the evolution of the price of a security and the associated volatility as a coupled system of SDEs. Such SDEs are commonly used as methods for pricing financial derivatives. Differing from our viewpoint, these models are typically used to price stock options under risk-neutral measures, with Volt and Magpie we are focused on performing predictive inference by conditioning on observations.

The connection between Gaussian processes and SDEs has been extensively studied by \citet{sarkka2019applied} who suggest Kalman filtering based approaches for estimating GP hyperparameters 
in SDE-inspired GP models, which we do not consider here, preferring simply marginal likelihood based estimation.
Systems of linear differential equations have been integrated into GP models previously via latent force models both for ordinary differential equations \citep{alvarez2009latent} and partial differential equations \citep{sarkka2011linear}.
To perform inference, \citet{alvarez2009latent} derive covariances corresponding to the linear projection of the differential operator onto a specific kernel, while \citet{sarkka2011linear,sarkka2019applied} use the projection operator explicitly to develop kernel functions to emulate systems of SDEs. 
Similarly, \citet{zhu2013locally} use SDEs to derive a nested GP, but their approach produces a standard GP with a non-deep, but structured, covariance function.
Autoregressive mean functions for GPs have been explored in \citet{gonzalvez2019financial}, however in their approach they use autoregressive features as inputs to a GP model, rather than as a way to specify the prior functions.

While many of the references above are focused specifically on finance, Volt and Magpie are applicable to a broad set of domains including climate modeling.
Autoregressive and volatility models have successfully been applied to domains such as wind and precipitation forecasting as in \citet{mehdizadeh2020estimating, liu2011comprehensive} and \citet{tian2018wind}.

The Gaussian process autoregressive model \citep{requeima2019gaussian} stacks Gaussian processes of different tasks, using the GP for one task as the mean function for the next.
It thus bears only slight resemblance to our moving average or multi-task approaches.
Furthermore, many well-studied autoregressive models, e.g. the AR(p) family, can be written as Gaussian processes \citep{williams_probabilistic_2010}.
As an alternative to developing domain specific kernel functions, one could alternatively construct manual combinations of generic kernels, which either requires significant amounts of hand-tuning as in \citet[][Ch 5.4,]{rasmussen_gaussian_2008} or solving discrete optimization problems \citep{lloyd2014automatic,sun2018differentiable}.
As we wish to develop our models efficiently and succinctly, we also do not consider these models.

\section{Methods}\label{sec: methods}

We first begin with a brief overview of Gaussian process regression models, before deriving the Volt kernel and Magpie mean functions in Section \ref{sec: derivation}. After deriving the Volt kernel and Magpie mean, we explain the inference procedure in Section \ref{sec: inference} and how we perform forecasting in Section \ref{sec: predictions}.

\paragraph{Gaussian Processes}
Please see \citet{rasmussen_gaussian_2008} for a more detailed introduction to Gaussian processes (GPs).
We assume noisy observations $y(t) \sim \mathcal{N}(f(t), \sigma^2),$ where $f \sim  \mathcal{GP}(\mu(t), k(t,t')),$
so that $\sigma$ is the observation noise and $f$ is drawn from a GP with mean function $\mu(t)$ and $k(t,t')$ as the covariance function.
When using GPs, we can compute the posterior predictive distribution, $p(f(\mathbf{t}^*) | \mathcal{D}),$ $\mathcal{D}:=\{\mathbf{t}, \mathbf{y}\}$, over new data points $\mathbf{t}^*$ 
is given by
$p(f(\mathbf{t}^*) | \mathcal{D}, \theta) = \mathcal{N}(\mu_{f | \mathcal{D}}^*, \Sigma_{f | \mathcal{D}}^*)$ where
	$\mu_{f | \mathcal{D}}^* = K_{\mathbf{t}^* \mathbf{t}} (K_{\mathbf{t}\mathbf{t}} + \sigma^2 I )^{-1} (\mathbf y - \mu(\mathbf{t})) + \mu(\mathbf{t}^*)$
	and 
	$\Sigma_{f | \mathcal{D}}^* = K_{\mathbf{t}^*\mathbf{t}^*} - K_{\mathbf{t}^* \mathbf{t}} (K_{\mathbf{t}\mathbf{t}} + \sigma^2 I )^{-1}K_{\mathbf{t} \mathbf{t}^*}$
with $K_{A, B}:=k(A, B)$.

\subsection{Volt and Magpie}\label{sec: derivation}

We make the common assumption that both the data $S(t)$, and volatility $V(t)$, have paths with log-normal marginal distributions. We therefore place the following joint SDE structure over $s(t) = \log S(t)$ and $v(t) = \log V(t)$,
\begin{equation}\label{eqn: joint-bm}
\begin{aligned}
ds(t) &= \mu_s dt + V(t)dW(t)\\
dv(t) &= -\frac{\sigma^2}{2}dt  + \sigma dZ(t).
\end{aligned}
\end{equation}
The drift term in Equation \eqref{eqn: joint-bm}, $ -\frac{\sigma^2}{2}dt$, arises from the log-transformation of the volatility, and ensures that forecast distributions over volatility have a constant mean (for further details see Appendix \ref{app: proofs}).
Furthermore, this structure allows us to derive closed form expressions for and auto-covariance functions associated with both log-data and log-volatility, allowing us to define the Volt model. 

Equation \eqref{eqn: joint-bm} gives a relationship between the log-price and log-volatility that is mirrored by many stochastic volatility models, including GARCH and SABR, where the volatility of the price is itself governed by an SDE \citep{bollerslev1986generalized,hagan2002managing}.
By recasting Equation \eqref{eqn: joint-bm} as a system of GPs we can move from an SDE sampling approach to a proper forecasting system based on historical observations.

\paragraph{A Gaussian Process Perspective}
Since for any finite collection of time points, $\mathbf{t} = \{t_i\}_{i=1}^{N}$, the observations $v = v(\mathbf{t})$ and $s = s(\mathbf{t})$ each have a multivariate normal distribution, $v$ and $s$ now correspond to Gaussian processes. Therefore we only need to derive the mean and covariance functions of the two processes to fully cast our problem as one of forming predictive distributions from GPs.

As $v(t)$ is a scaled Wiener process with constant drift term, the autocovariance function is 
\begin{align}
    K_v(t, t') = \sigma^2 \min\left\{t, t'\right\}
    \label{eq:scaled_bm}
\end{align} and the mean is $\mu_v(t) = -t\frac{\sigma^2}{2} $ so that, $v(t) \sim \mathcal{GP}\left(\mu_v(t), K_v(t, t')\right)$.
Conditional on a realization of $V(t) = \exp v(t)$, $s(t)$ is also described by a Gaussian process with $\mathbb{E}[s(t)] = \int_{0}^{t}\mu_s dt = t \mu_s$ and  ,
\begin{align}
 Cov(s(t), s(t')) = \int_{0}^{\min\{t,t'\}}V(t)^2 dt = K_s(t, t'; V(t)),
 \label{eq:vol_kernel}
\end{align}
producing our model over log-data:
\begin{equation}\label{eqn: price-gp}
    s(t) \sim \mathcal{GP}\left(t\mu_s + s(0), K_s(t, t'; V(t))\right).
\end{equation}

The final Volt model is then a hierarchical composition of Gaussian processes:
\begin{equation}\label{eqn: full-model}
\begin{aligned}
   v(t) &\sim \mathcal{GP}(m_v(t), K_v(t, t'))\\
   V(t) &= \exp\left(v(t)\right)\\
   s(t) &\sim \mathcal{GP}(m_s(t), K_s(t, t'; V(t)))\\
   S(t) &= \exp\left( s(t) \right),
\end{aligned}
\end{equation}
The log-volatility is distributed as a Gaussian process dependent on the the time inputs, the mean $m_v$, and the \emph{volvol} hyperparameter $\sigma$ and has a Brownian motion covariance (Eq. \ref{eq:scaled_bm}).
Given a realization of a volatility path over time and the parameters of the log-linear mean, the log-price is also distributed as a Gaussian process with covariance given by Eq. \ref{eq:vol_kernel}. To generate predictions using the log-volatility and log-price GPs we first must infer both a volatility path 
from the observed time series, $S = S(\mathbf{t})$, and the hyperparameters of both the data and volatility models. A complete derivation of the GPs in Equation \eqref{eqn: full-model} is in Appendix \ref{app: proofs}.

\paragraph{Magpie} 
For the sake of deriving the covariance functions associated with the log-data and log-volatility processes, we have left the mean functions of the data GP in Equation \eqref{eqn: full-model} as a simple linear function. While we may believe that there are nontrivial trends in the data over time, we also believe that these trends may be more complex than simple polynomial or periodic functions, and in the context of applications like finance and climatology are likely to change over time with evolving market or climatological conditions. 

To address these deficiencies in using simple mean functions in modeling nonstationary signals we replace the simple mean functions typically found in GP models with exponential moving averages (EMA) 
\citep{nau2014forecasting}. We use the EMA with a limited number of terms, defined as
\begin{equation}\label{eqn: ema}
    \begin{aligned}
    EMA(\mathbf{s})_{i+1} = &\alpha[s_{i} + (1-\alpha)s_{i-1} + (1-\alpha)^2 s_{i-2} \\
    &+ \cdots + (1-\alpha)^{k-1}s_{i-(k-1)}]
    \end{aligned}
\end{equation}
where $\alpha = 2/(k + 1)$ is a hyperparameter governing the smoothing of the moving average. 
A smaller value of $k$ uses only more recent observations, enabling a closer match of the data, whereas a larger value of $k$ uses more data and smooths the data more.

While we focus on the EMA in Equation \eqref{eqn: ema}, Magpie naturally extends to alternate moving averages, such as lag-corrected moving averages. We provide comparisons of these alternate moving averages, as well as the effect of the $k$ hyperparamter in Appendix Figure \ref{fig: moving-average-example} and in the extended results of Section \ref{sec:forecasting}. With Equation \eqref{eqn: ema} we can define the Magpie mean function as $m_{EMA}(t_{i+1}, \mathbf{s}) = EMA(\mathbf{s})_{i+1}$.

We close this section by noting that moving from a linear to a exponential moving average mean for the GPs breaks the connection with the SDEs described in Section \ref{sec: derivation}, making the combination of Volt and Magpie necessarily a practical approach, rather than an entirely theoretically motivated approach.

\subsection{Inference}\label{sec: inference}

Here we outline the procedure for using a series of price observations to train the hyperparameters of the GPs in Equation \ref{eqn: full-model}, and form the associated posterior predictive distributions. In general the training procedure can be thought of as a three step process: 
a) use a Gaussian Process Copula Volatility (GPCV) model to infer a volatility path, $V$, given a sequence of observations $S$, b) learn the hyperparameters of the GP in log-volatility space by maximizing the Marginal Log-Likelihood (MLL) with respect to the GPCV inferred volatility, c) learn the hyperparameters of the GP in log-data space by maximizing the MLL with respect to the observed prices, using the kernel generated by the GPCV inferred volatility path.
Note that our use of the GPCV to estimate volatility is a modelling choice and we could alternatively have used any other volatility estimation model such as GARCH.

\paragraph{Inferring Volatility from Training Data} One challenge in formulating the model outlined in Equation \eqref{eqn: full-model} is the need to have both data and volatility observations for some range of training observations. 
To estimate the volatility, we use a variant of Gaussian copula process volatility (GPCV) model first proposed by \citet{wilson_copula_2010}.
Our GPCV model uses a warped Gaussian process to model the variability of the responses, $w(t)$, according to:
\begin{equation}
\begin{aligned}
    f(t) &\sim \mathcal{GP}(c, K_v(t,t')) \\
    \gamma(f(t)) &= \exp\{f(t)\}\\
    w(t) &\sim \mathcal{N}(0, \gamma^2(f(t))).
\end{aligned}
\end{equation}
We use the kernel derived from log-volatility SDE in Equation \eqref{eqn: full-model} to infer the latent function $f(t),$ and use variational inference \citep{hensman2013gaussian,hensman2015scalable} to train the model. 
See Appendix \ref{app:gpcv_training} for further details.

Following \citet{wilson_copula_2010}, we consider the responses as the log-returns of the data, that is: $w(t_i) = \log S(t_i) - \log S(t_{i - 1}).$
We construct a volatility prediction over times $0, \cdots, t-1$ by drawing posterior samples from $f(t)$ and passing them through the warping function $\sigma(\cdot);$ so our estimate for $V(t)$ is 
\begin{equation}
    \hat V(t) := \frac{1}{J} \sum_{j=1}^J \gamma(f_j(t)), \hspace{0.1cm} f_j(t) \sim q(f(t) | w(t), v, \theta), \label{eq:vol_sampling}
\end{equation}
where $q(f(t) | w(t), v, \theta)$ is our approximate posterior distribution over the latent function $f(t).$
We demonstrate that our approach is able to correctly estimate the true volatility in Figure \ref{fig: intro}, where the volatility and price are drawn from a SABR volatility model \citep{hagan2002managing}.

\paragraph{Training the Gaussian Processes}
Given the volatility path associated with the training data learned using a GPCV, we assume a Gaussian process priors over the log-volatility and log-data according to Equation \eqref{eqn: full-model}.
Given the volatility over the training data, the single hyperparameter of the log-volatility model is the $\sigma^2$ term describing the \emph{volvol}. The hyperparameters of the log-data model are just the parameters of the mean in Equation \eqref{eqn: full-model}, of which there are none if we are using a non-parametric mean like Magpie.
To train we maximize the MLL of the models with respect to their hyperparameters using gradient based optimization \citep[Chapter 5]{rasmussen_gaussian_2008}.
The total computational cost for inference in Volt, regardless of the use of a Magpie mean, is just the cost of training \emph{one} variational GP and two standard GP models on evenly spaced data, which can be done efficiently via exploiting the (Toeplitz) structure of the data \citep{wilson2015kernel}. 

\begin{figure*}[h!]
    \centering
    \includegraphics[width=0.9\linewidth]{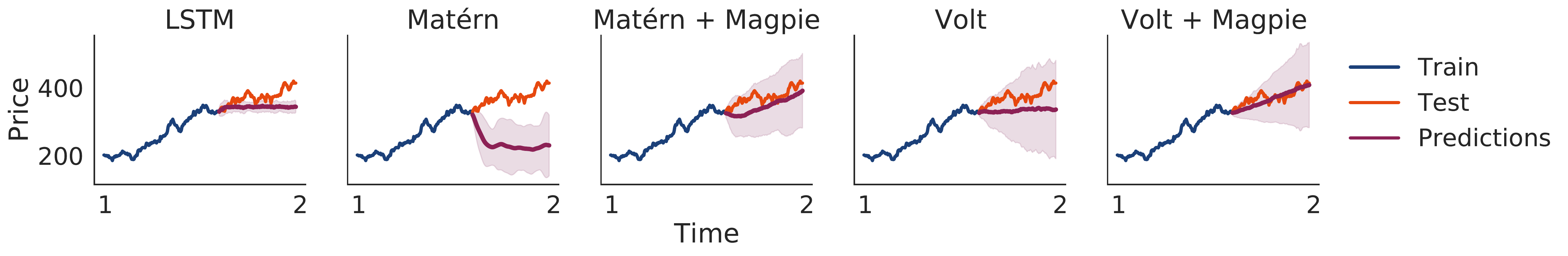}
    \caption{Simulations and forecasts showing the mean and $95\%$ confidence region for various model choices. Probabilistic LSTMs perform well on the training data, but do not extrapolate far from observed data well. Mat\'ern forecasts quickly revert to constant level of uncertainty which leads to overconfidence far away from observations, whereas Volt's increase in uncertainty as we move away from training data produces well calibrated to the data. The constant mean forecasts in both Mat\'ern and Volt fail to pick up the long term trend in the data, which Magpie means accurately capture.
    The combination of Volt and Magpie, with correct inductive biases in both the kernel \emph{and} mean functions produces forecasts consistent with trends in the data and with well calibrated uncertainty.}   
    \label{fig:sample-rollouts}
\end{figure*}

\subsection{Predictions}\label{sec: predictions} 
In Volt, we condition the log-volatility GP on a log-path inferred by GPCV and the log-data GP on historical observations of log-price and draw samples from the \emph{posterior} distributions, producing a mixture of log-normal distributions over data. 
Sampling the posterior requires sample $N_v$ log-volatility paths, $v^*$, over the test inputs and for each of these we generate a kernel $K_s(t, t', V^*)$, and sample $N_s$ data paths, $S^* = \exp(s^*),$ producing $N_v \times N_s$ samples.

In the Gaussian process viewpoint of Section \ref{sec: derivation}, standard Monte Carlo simulation of an SDE procedure is equivalent to sampling log-volatility paths, $v^* = v(\mathbf{t}^*),$ from the \emph{prior} distributions of Equation \eqref{eqn: full-model} up to time $T$ rather than the \emph{posterior} distribution \citep{sauer2012numerical}.
With the prior samples of $S_T^* = \exp(s_T^*),$ we can form a Monte Carlo estimate of future distributions over price. However, the distinction between this type of approach and our approach for sampling with Volt is that Volt samples from the \emph{posterior} distributions over volatility and data conditional on observations, while the SDE based approaches sample from the prior distribution over volatility.

\paragraph{Rollout Predictions} The Magpie mean only allows for predictions one step ahead, so we do our forecasting in a \emph{rollout} fashion. That is, we use observations $s_0, \dots, s_t$ to sample $\hat{s}_{t+1}$ from the GP posterior $p(s_{t+1} | s_0, \dots, s_t)$, then condition our GP (and Magpie mean) on $\hat{s}_{t+1}$ in order to sample $\hat{s}_{t+2}$ from the updated GP posterior $p(s_{t+2} | s_0, \dots, s_t, \hat{s}_{t+1})$, and so on. 
These rollout forecasts are critical to the Magpie framework. By sequentially sampling the price forecasts and updating the GP with each observation we allow for trend reversals in the moving average mean in a way that is not possible with other GPs.
Rollouts are unnecessary for traditional means because the conditional means over each time step factorize into a single multivariate Gaussian distribution.

\section{Forecasting}\label{sec:forecasting}

In both financial and climatological applications we are considering the data as stochastically evolving and are thus interested in forecasting \emph{distributions} over outcomes, rather than point estimates.
For this reason we use \emph{calibration} and negative log likelihood as our primary measures of interest, rather than an accuracy metric like mean squared error.

\begin{figure}[ht!]
    \centering
    \includegraphics[width=\linewidth]{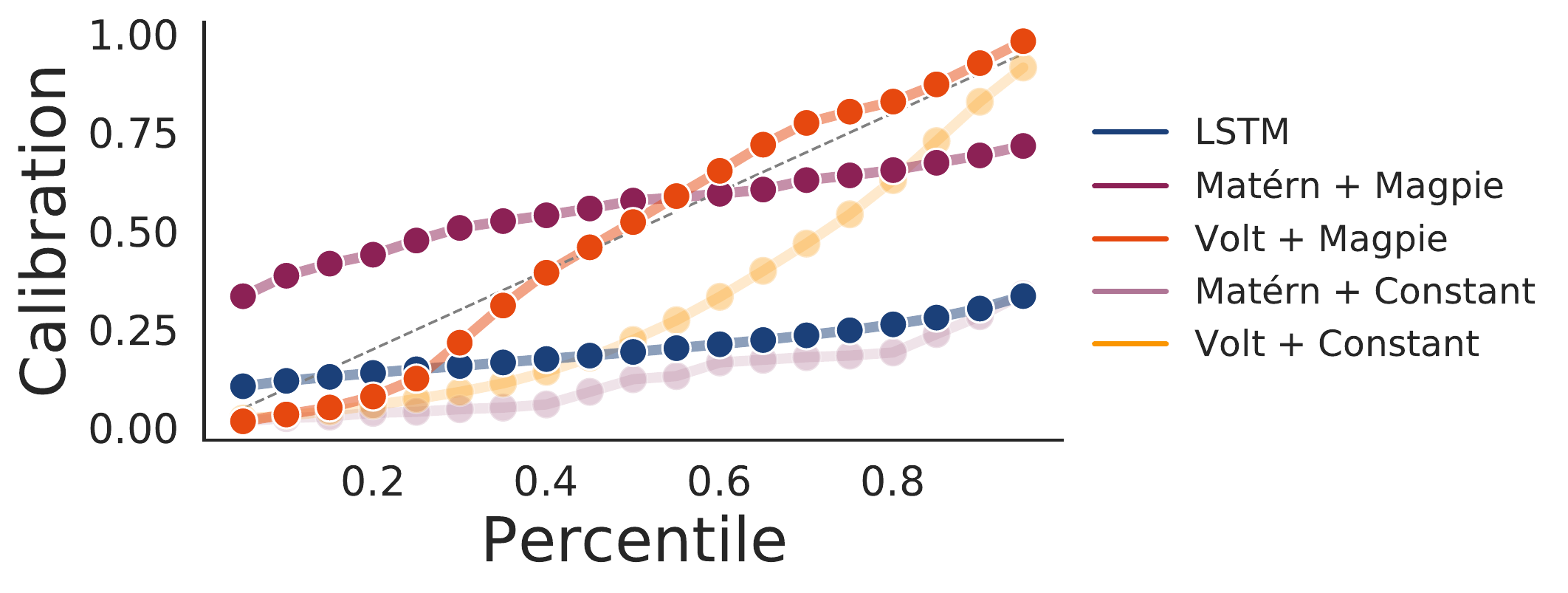}
    \caption{Calibration of various approaches on the $2$ years of data from the NASDAQ $100$. The forecasts generated by standard kernels and probabilistic LSTMs are significantly overconfident, leading to very poor calibration.}
    \label{fig:stock-calib}
\end{figure}

We compute the calibration at percentile $p$ by computing the frequency with which the true observation is less than the empirically computed $p^{th}$ quantile of the forecast distribution. 
More specifically, for a forecast of the price stock $S$ at time $T$ and percentile $p$ we compute the empirical quantile of the forecast $q_{T}$ where $\hat{\mathbb{P}}(S_{T} < q_{T}) = p$. We can then compute the calibration at $p$ as the empirically observed frequency of the event $S_{T} < q_{T}$ by calculating $C_p =\frac{1}{K}\sum_{k}\mathbb{I}_{\{S_{T_k} < q_{T_k}\}}$
as the average frequency of $S_T < q_T$ over $K$ different forecasts.
If our forecasts are well calibrated then this empirical frequency will be close to $p$ for each value of $p$; therefore by computing the calibration of our forecasts at a range of percentiles, $p$, we can determine the overall calibration of the forecast distribution.  Such a calibration metric is similar to those explored for regression in \citet{kuleshov2018accurate}.

Note that for accurate calibration to occur in this setting our forecast distribution must match the empirical observations at all quantiles. We could not, for example, just forecast that a price increases some fixed percentage of the time that matches the observed frequency of the price increased and expect to achieve accurate calibration.

\subsection{Stock Price and Foreign Exchange Rate Forecasting}\label{sec:stocks}

As Volt and Magpie are primarily inspired by financial time series models, forecasting distributions over stock prices is a core application of our approach. We compare Volt and Magpie to baseline models of GPs with standard kernel and mean functions. Along with these GP models, we include probabilistic LSTMs where we optimize a predicted mean and variance at each time step with respect to the negative log-likelihood (NLL) which have been previously used in a quantitative finance setting \citep{chauhan2020uncertainty}. All models assume the marginal distributions of the observations are normally distributed, thus we model the log-price of stocks in each case. 

Figure \ref{fig:sample-rollouts} provides a representative comparison of forecasts generated by GPs both with and without Volt and Magpie, and the probabilistic LSTMs used here. Simpler probabilistic models like standard GPs and probabilistic neural networks generally provide overconfident forecasts, and more traditional mean functions in GP models do not capture the long range trends that are commonly present in financial time-series data.

Figure \ref{fig:stock-calib} shows the calibration of the compared methods aggregated over thousands of forecasts.
We consider stocks in the NASDAQ $100$ collection, and a history of two years of daily observations leading up to January $2022$. For $25$ evenly spaced days we forecast $1000$ paths $100$ days into the future and compute the calibration curves of the forecasts for the days $75$ to $100$ days out.

We see in Figure \ref{fig:stock-calib} that Volt is able to remedy a significant overconfidence that is present in alternative methods such as standard GP kernels or probabilistic LSTMs. Furthermore, it is the Magpie mean function that enables the distributions to be centered at the correct values, which is why we see the LSTMs and constant mean GPs showing bias in the calibration plots.
Note that for accurate calibration to occur in this setting our forecast distribution must match the empirical observations at \emph{all} quantiles. We could not, for example, just forecast that a price increases some fixed percentage of the time that matches the observed frequency of the price increased and expect to achieve accurate calibration.
\begin{table}[h!]
\begin{small}
    \centering
    \begin{tabular}{c|c c}
         & Stock Prices & Wind Speeds\\
         \hline
         Volt + Magpie & $5.88 \pm 0.02$ & $4.28 \pm 0.16$\\
         
         Volt + Con. &  $\mathbf{4.69 \pm 0.03}$ & $\mathbf{3.38 \pm 0.05}$ \\
         
         Mat\'ern + Magpie & $9.80 \pm 0.27$ & $12.13 \pm 0.81$\\
         
         Mat\'ern + Con. & $7.74 \pm 0.21$ & $18.03 \pm 1.90$\\
         
         SM + Magpie & $147.84 \pm 1.84$ & $110.07\pm 7.81$\\
         
         SM + Con. & $80.43 \pm 0.57$ & $70.14\pm 5.03$\\
         
         LSTM & $49.95 \pm 0.59$ & $45.13\pm 1.82$\\
         
         Volt-VHGP + Con. & $4.76 \pm 3.05$ & $5.75 \pm 0.44$\\
         
         Volt-VHGP + Magpie& $6.97 \pm 1.24$ & $5.91 \pm 0.34$\\
         
         GPCV & $5.45 \pm 1.51$ & $4.89 \pm 0.04$\\
         \bottomrule
    \end{tabular}
    \caption{Negative log likelihoods (NLLs) per test point with $2$ standard deviations for the methods compared on both the stock forecasting and wind speed tasks. By accounting for uncertainty in both the volatility and the data forecasts, Volt provides highly accurate test distributions relative to baseline approaches. Volt-VHGP indicates a Volt model where we use variational heteroscedastic GPs from \citet{lazaro2011variational} in place of GPCV. We provide expanded results including foreign exchange data in Appendix \ref{app: exps}. In each case the mean and standard deviation are computed over approximately $2$ thousand time series $75$ to $100$ time steps into the future, yielding tens of thousands of individual forecasts.}
    \label{tab:nlls}
\end{small}
\end{table}

Table \ref{tab:nlls} gives the average test negative log likelihood (NLL) values on the stock forecasting task and on the foreign exchange data from \citet{lai2018modeling}. Both variants of the Volt model outperform competing methods such as LSTMs, and GPs with Mat\'ern and Spectral Mixture (SM) kernels \citep{wilson2015kernel}. Volt with a constant mean is slightly better than with a Magpie mean in terms of NLL, the Magpie mean is key to achieving high calibration, as we see in Figure \ref{fig:stock-calib}. SM kernels are a highly class of flexible kernel, but rely on there being frequency components in the data, with non-stationary data such as those studied here, the lack of regularity in the data leads to weak performance.

\subsection{Wind Speed Forecasting \label{sec: wind}}

Probabilistic forecasting models play an important role in statistical climatology in providing forecast distributions over quantities of interest, such as rainfall or wind speed, that can be used to generate synthetic data or estimate the risk of extreme events. Stochastic volatility models have a history of use in modeling wind speed, but typically these models have been limited to GARCH based approaches \citep{liu2011comprehensive,tian2018wind}, and are thus focused on the \emph{volatility} of wind, rather than forecasting distributions of wind speed itself. 

\begin{figure}
    \centering
    \includegraphics[width=\linewidth]{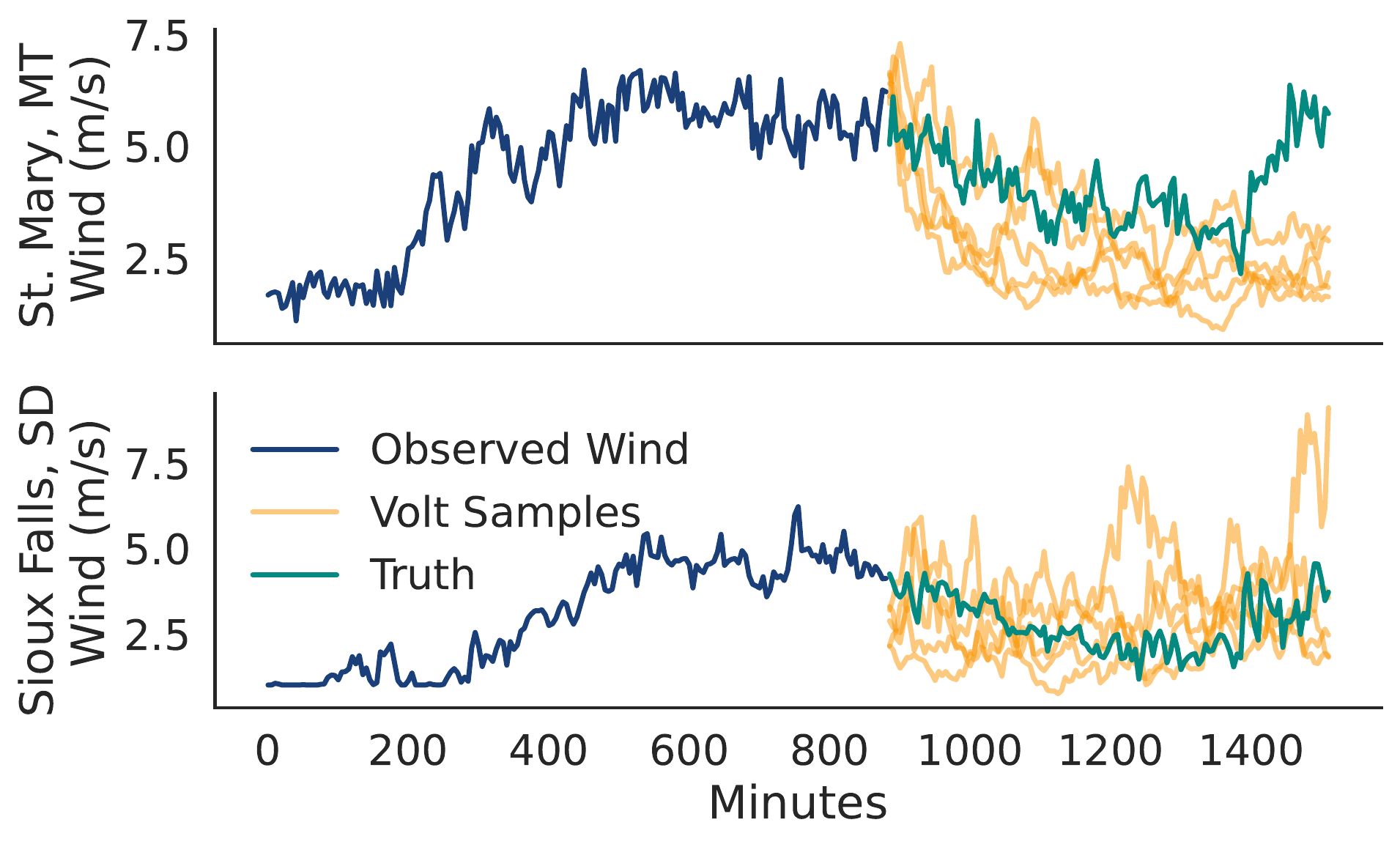}
    \caption{A representative example of observed wind speed and samples of multitask Volt forecasts for two related observation stations. While none of the Volt forecasts perfectly fit the true future observations of wind, each individual roll is a realistic potential realization of wind speed. By generating many plausible outcomes we are able to forecast distributions over wind speed that are highly calibrated to held out test observations.}
    \label{fig:wind-example}
\end{figure}

Here we apply Volt and Magpie to the problem of developing a stochastic weather model for wind speed. We source historical wind data from the U.S. Climate Reference Network (USCRN), with observations taken at $5$ minute intervals over the $2021$ calendar year at $154$ spatial locations in the United States \citep{diamond2013us}. Figure \ref{fig:wind-example} provides an example of what the wind speed observations look like, as well as example Volt forecasts in comparison to ground truth held out data.
Each forecast path represents a realistic scenario drawn from a distribution over paths from which the true data would, hypothetically, be a representative candidate.
By sampling paths from the forecast distribution over wind speeds we can simulate future observations with accurate probability enabling us to estimate statistics of interest, such as expected wind speed or the probability of extreme events. 

\begin{figure}
	\centering
    \includegraphics[width=1.\linewidth]{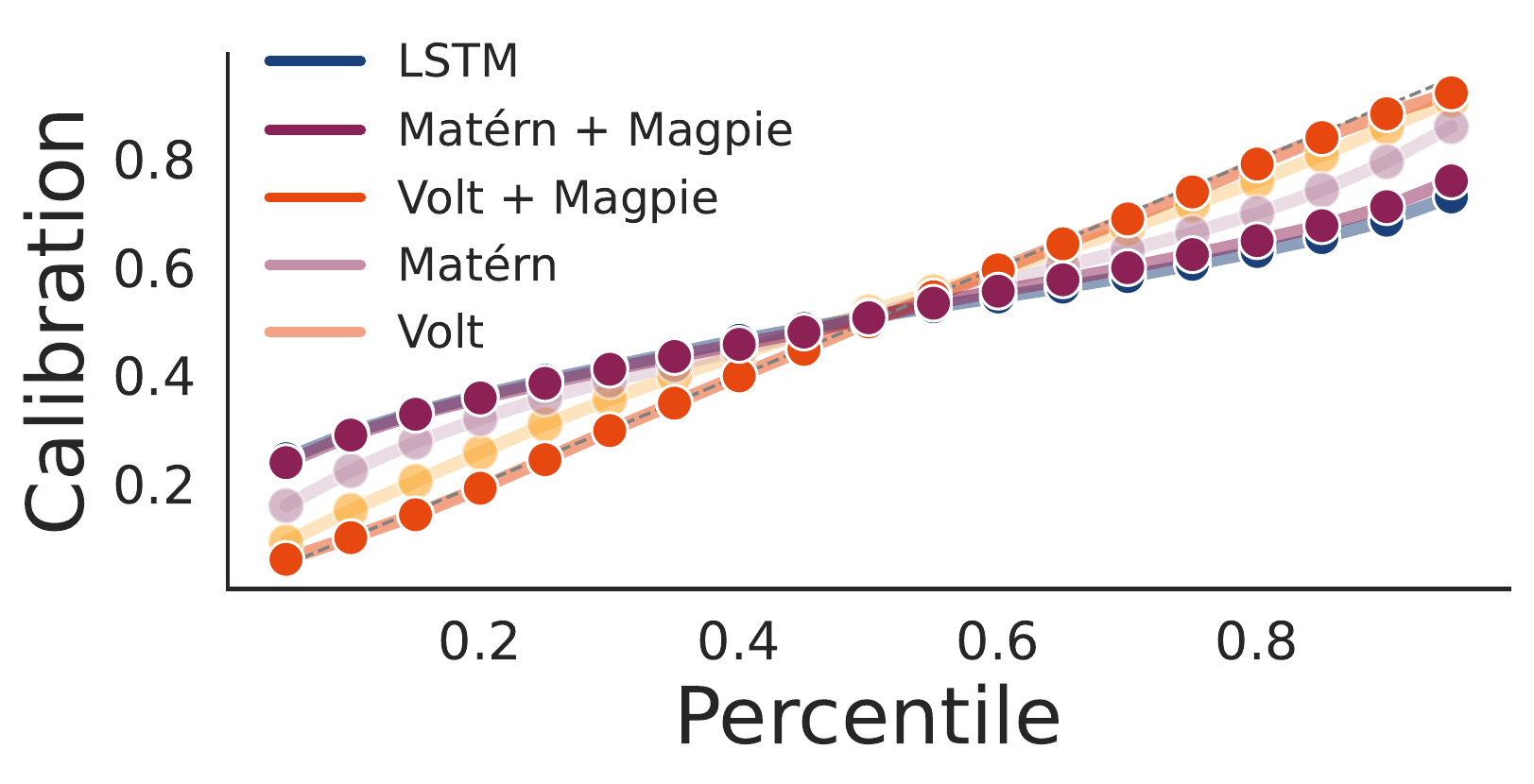}
    \caption{A calibration plot for wind speed, aggregated over hundreds of thousands of distinct forecasts.
    While probabilistic LSTMs and standard GP models provide competitive baselines, the Volt and Magpie model generates wind speed distribution forecasts that are extremely well calibrated. These calibrated distributions enable us to quickly simulate thousands of scenarios that can be trusted to faithfully represent potential outcomes.}
    \label{fig:wind-calib}
\end{figure}

As with stock price forecasting, we are interested in producing forecast distributions that match the ground truth of the data, rather than attempting to generate point predictions.
In Figure \ref{fig:wind-calib} we compare the calibration of forecasts against the ground truth wind speeds in the forecast windows. Table \ref{tab:nlls} provides the NLL values of the various approaches. As with stock forecasting, we see that constant means do provide slightly better NLL values than Magpie means, but Volt models are key to producing accurate forecasts. 
Distinct from stock price forecasting however, is the bounded nature of wind speed. As we do not expect wind speed to grow indefinitely (as we may see with stocks) we forecast with a small amount of mean reversion applied to the GP models.  
Experimental details, including a sensitivity to the mean reversion can be found in Appendix \ref{app: wind}.

\section{Multi-Task Volatility Modelling}\label{sec: multitask}

Finally, we extend Volt to model several asset prices at once by using multi-task Gaussian processes with the goal of jointly modelling different time series at once, such as the wind speeds for the continental United States.

First, we extend the GPCV of \citet{wilson_copula_2010} to several tasks before then placing a multi-task model over volatility in the hierarchical GP formulation.
Our approach enables simultaneous estimation of the time series, its volatility, and the relationships between the time series themselves.
\begin{figure*}[t!]
	\centering
	\begin{subfigure}{0.42\textwidth}
	    \centering
	    \includegraphics[width=\linewidth]{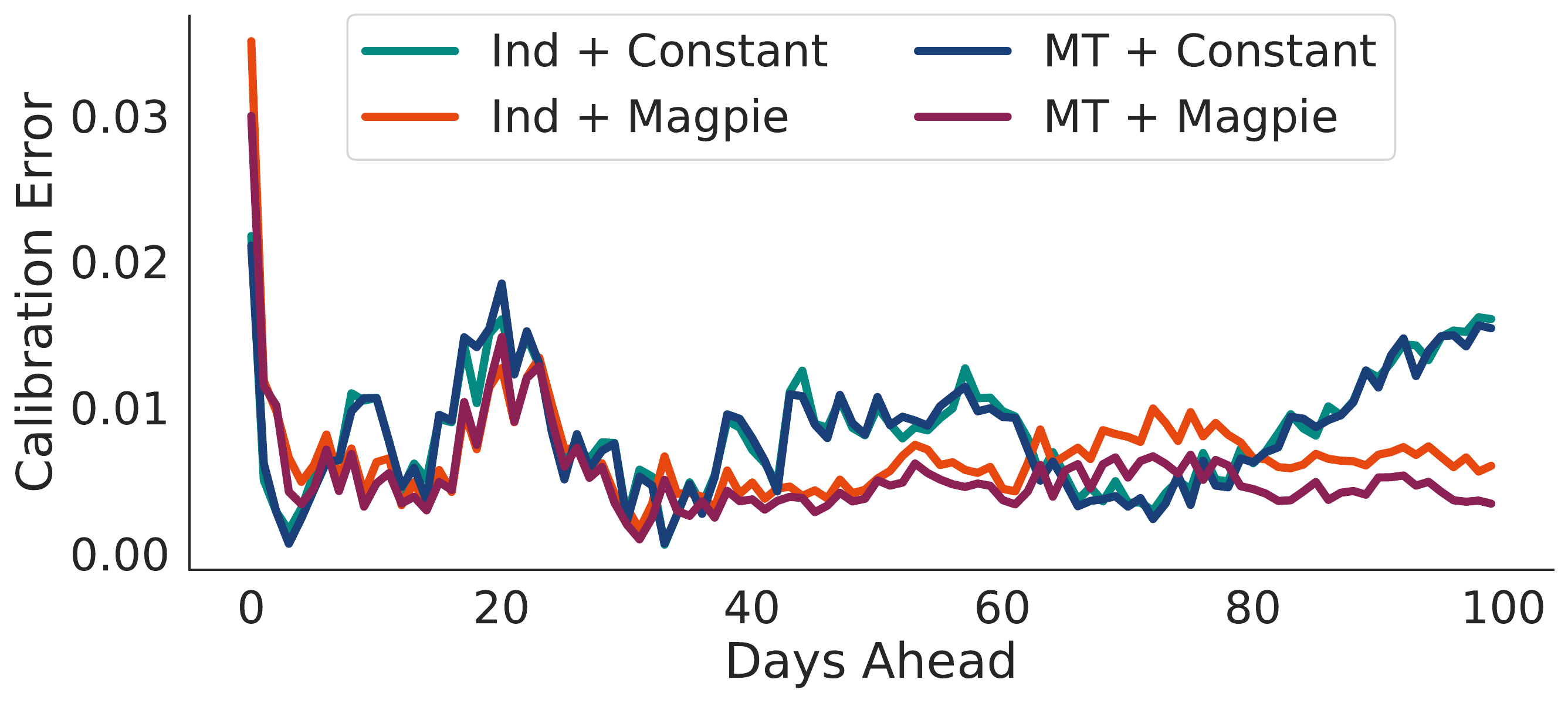}
	\end{subfigure}
	\begin{subfigure}{0.25\textwidth}
	    \centering
	    \includegraphics[width=0.9\linewidth]{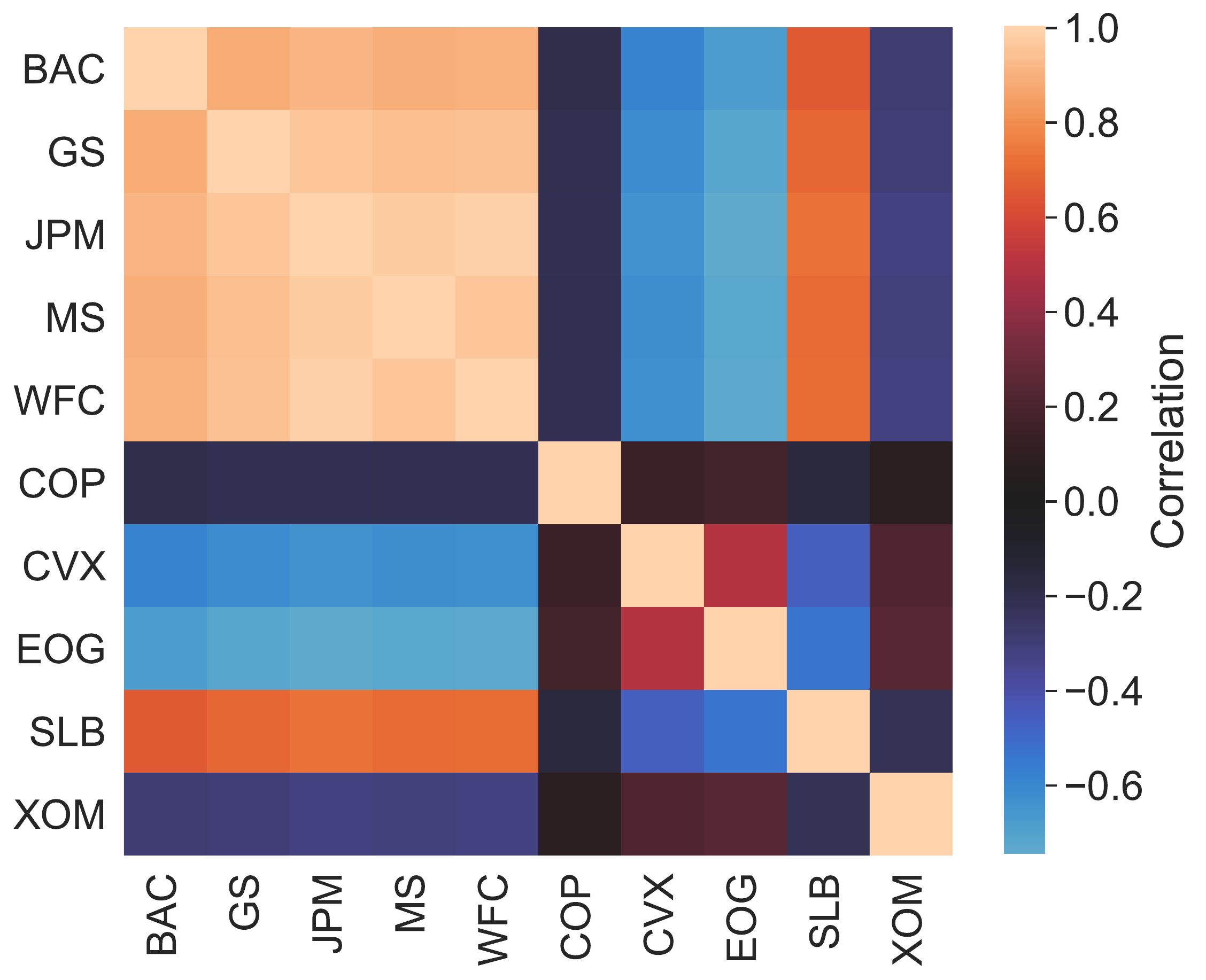}
	\end{subfigure}
	\begin{subfigure}{0.3\textwidth}
	    \centering
	    \includegraphics[width=\linewidth]{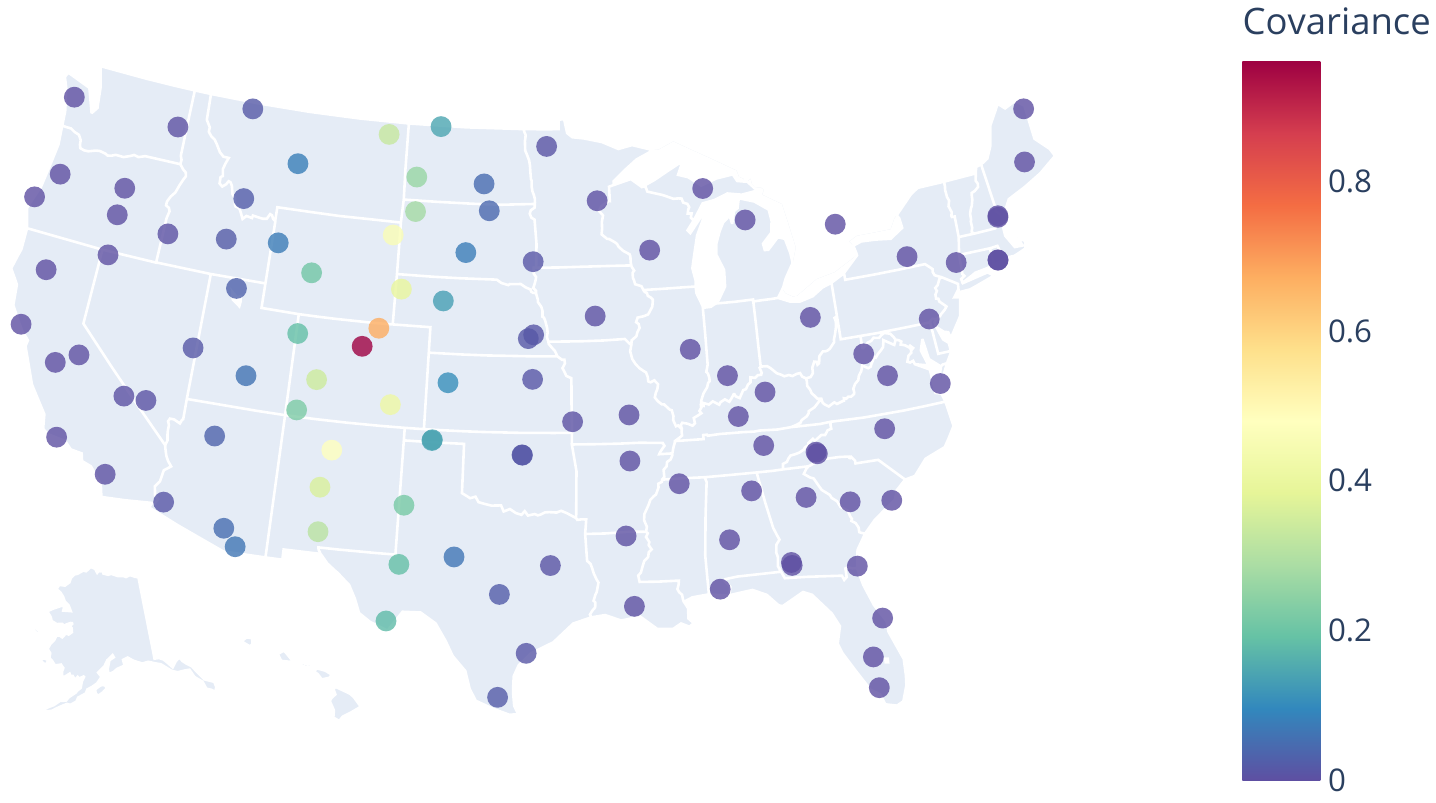}
	\end{subfigure}

    \caption{
    \textbf{Left panel:} Calibration error of both Volt and MT Volt at the $95\%$ confidence level as a function of time step lookahead over $30$ stocks from an entire sector ETF (XLF). While both are well-calibrated, MT Volt preserves well-calibration to longer time steps.
    \textbf{Center panel:} Estimated correlation matrix of stocks from two different sectors. MT-Volt successfully learns the high volatility correlation amongst the finance sector stocks (first five) with lower correlations between the energy sector stocks (second five).
    \textbf{Right panel:} Estimated volatility covariance with Boulder, CO. Correlations decrease as the stations go further away.
    }
    \label{fig:mt_volt_experiments}
\end{figure*}

\subsection{A Multitask GPCV}
We extend the Gaussian process copula volatility model (GPCV) described in Section \ref{sec: inference} to model several jointly related volatilities at once by using multi-task Gaussian processes \citep{bonilla2007multi,alvarez2012kernels}.
We assume that all returns and volatilities are observed at once, with $P$ different responses, so that the covariance between the $p$th and $p'$th latent Gaussian process is given by: $k([t, p], [t', p']) = K_v(t,t')K_i(p, p')$ where $K_i(p, p')$ is a lookup table describing the intertask covariance.
The intertask covariance is a $P \times P$ matrix; we can regularize it with a LKJ prior \citep{lewandowski2009generating} or to incorporate side information such as geographic coordinates.

We have the multi-task probabilistic model:
\begin{equation}
\begin{aligned}
    f_p(t) &\sim \mathcal{GP}(c, K_v(t,t') K_i(p,p')) \\
    w_p(t) &\sim \mathcal{N}(0, \exp^2(f_p(t)). 
\end{aligned}
\end{equation}
Again, we use variational inference to infer the latent posterior distribution over each price's latent Gaussian process, see Appendix \ref{app:gpcv_training} for more details.
We also use posterior samples from this multi-task GPCV to estimate volatility, $\hat{V}_p(t),$ for each stock price $p$ by following Eq. \ref{eq:vol_sampling}.

In Appendix Figure \ref{fig:mt_gpcv}, we simulate price data from a correlated SABR volatility model and use our multi-task GPCV to recover both the volatility as well as the latent correlation structures.
This suggests that our inference scheme enables us to accurately recover latent correlations.

\subsection{Multi-Task Stock Modeling}

After using a multi-task GPCV to estimate the volatility for us, we then use a multi-task Gaussian process model \citep{bonilla2007multi} to estimate volatility, producing the following probabilistic model:
\begin{equation}\label{eqn: mt-model}
\begin{aligned}
   v_p(t) &\sim \mathcal{GP}(m_v, K_v(t, t') K_p(t,t'))\\
   V_p(t) &= \exp\left(v_p(t)\right)\\
   s_p(t) &\sim \mathcal{GP}(m_{s,p}(t), K_{s,p}(t, t'; V_p(t)))\\
   S_p(t) &= \exp\left( s_p(t) \right).
\end{aligned}
\end{equation}
Conditional on the correlated volatility paths, the prices themselves are independent, so we use $P$ independent Gaussian process models to model the prices.
Intuitively, this dependency structure makes sense as we expect exogenous shocks (for example, large scale macroeconomic trends) to affect variability in an asset prices, rather than just directly producing an increase or decrease.

\subsection{Multitask Stock Price Prediction}

In Figure \ref{fig:mt_volt_experiments} left panel, we consider the calibration of both Volt and MT Volt on predictions across five different groupings of stocks each with between $5$ and $30$ different stocks in each group, finding that all models are fairly well calibrated in terms of the calibration error, which is the squared error of the average calibration across bins of the empirical observed calibration of the foreast \citep{kuleshov2018accurate}. The mult-task models tend to improve calibration over independent models, especially when using Magpie means.
We display the results for calibration across time steps, mean absolute error (MAE) and negative log likelihood (NLL) in Appendix \ref{app:mt_volt_experiments}.

In Figure \ref{fig:mt_volt_experiments} center, we showcase how the multi-task Volt model of volatility can be used to measure the relationships between assets.
We considered $10$ stocks, five from the financial sector and five from the energy sector.
Volt learns strong correlations amongst the stocks in the financial sector and much weaker cross-correlations with the energy stocks. 

\subsection{Spatiotemporal Wind Modelling}

Finally, we consider multi-task modelling for stochastic weather generation.
Here, as we have longitude and latitude coordinates for each of the weather locations, we can incorporate this information into the inter-task covariance matrix by using a geodesic exponential kernel, which is given as $k(x, y) = \exp\{-\arccos(x^\top y)/2\sigma^2\}$ for $x, y \in \mathbb{S}^2,$ that is points on the unit sphere \citep{jayasumana2013combining}.
Note that we have no restrictions on kernel choice and could alternatively consider non-stationary kernels here instead.

We model $110$ stations across the United States in the year $2021$ again at $5$ minute intervals, estimating the relationship between each station using the geodesic exponential kernel described above, and learning the lengthscale.
We display the results in Figure \ref{fig:mt_volt_experiments} right panel with the stations described on a map of the United States.
Further experimental results are shown in Appendix \ref{app:mt_volt_experiments}.

\section{Conclusion}
We have proposed Volt and Magpie, which use kernels and mean functions derived from stochastic volatility, in order to introduce a powerful forecasting method for stochastically generated time series.
Volt deviates from the usual assumptions of stochastic differential equation (SDE) models for financial and climatological models, and incorporates historical data through GPs, allowing us to better estimate expectations and forecast distributions. Magpie allows us to replace the often over-simplified mean functions in Gaussian process models with a nonstationary mean leading to forecasts that more closely represent the data.

We have demonstrated that Volt and Magpie can outperform competing methods in generating forecast distributions of stochastically generated processes, with an emphasis on financial and climatological applications.
The strong predictive uncertainties allow our method to be used for price forecasting and weather generation in a reliable way that produces trustworthy forecasts that are well calibrated to observations.
Finally, we proposed a multi-task extension to Volt that improves on Volt's predictive calibration while additionally allowing for the estimation of the relationships between several assets at once.

The potential applications of our approach are broad, with potential uses in financial domains such as automated trading and strategy development, and climatological research in which Volt and Magpie could serve as a backbone for large spatiotemporal climate models.
In the future, it would be useful to extend both the single and multi-task models to use online variational inference \citep{bui2017streaming,maddox2021conditioning} to enable online deployment of scalable forecasting strategies.
We hope our work will catalyze further development of domain based prior kernels for Gaussian processes, and applications of probabilistic machine learning to financial climatological data.

\section*{Acknowledgements}
We would like to thank Andres Potapczynski for helpful discussions. This research is supported by an Amazon Research Award, Facebook
Research, Google Research, Capital One, NSF CAREER IIS-2145492, NSF I-DISRE 193471, NIH R01DA048764-01A1, NSF IIS-1910266, and NSF 1922658 NRT-HDR.

\bibliographystyle{apalike}
\nocite{*}
\bibliography{refs}
\clearpage
\onecolumn

\appendix

\section{Tutorial}\label{app: tutorial}

This section should serve as a useful reference on much of the more domain-specific language and methodology used throughout the paper.

\subsection{Volatility}\label{app: tutorial}

In the context of a time series $S_t$, we use \emph{volatility}, denoted $V_t$, to refer to the standard deviation of the variability in price over some time period. In financial applications we consider stock prices on the daily time scale, and as is standard report volatility as \emph{annualized volatility}, which corresponds to the volatility of a stock over the course of a year. 

More specifically, we assume that the \emph{log returns} in observations, $\log\left(\frac{S_{t+1}}{S_t}\right)$, are normally distributed with standard deviation $V_{t}$. 
In this paper we make the common assumption that the volatility itself is a time varying stochastic process, meaning we expect the magnitude of the daily returns to vary over time. 

\begin{figure}[h!]
    \centering
    \includegraphics[width=\linewidth]{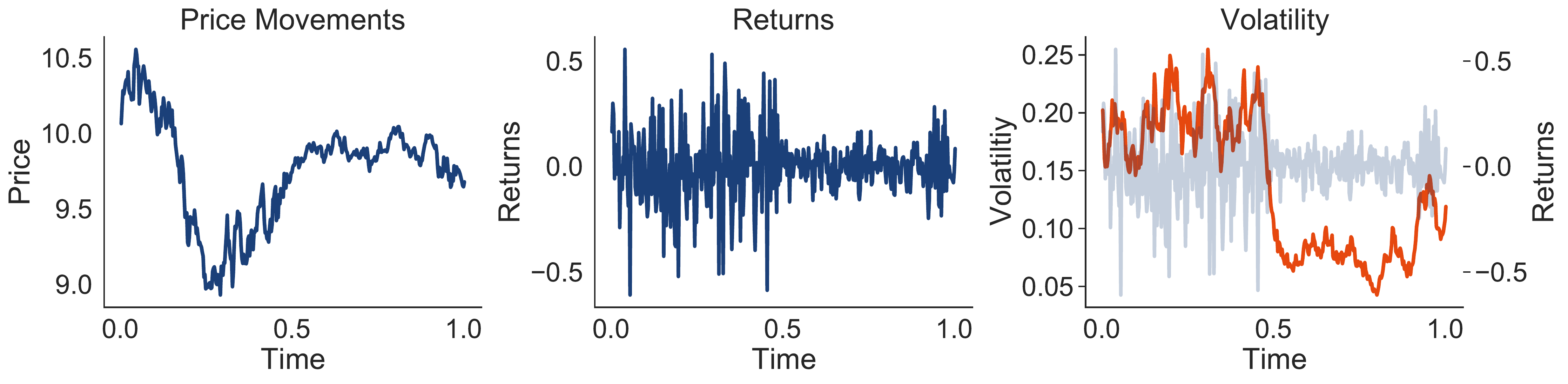}
    \caption{\textbf{Left: } Price movements over time. Movements tend to be larger up to time $t=0.5$ \textbf{Center: } Returns over time, as calculated by $S_{t+1}/S_t$. Here, returns are clearly larger in the first half of the time series. \textbf{Right:} Volatility overlaid with returns for the same price. Volatility is clearly higher when the returns have a larger absolute magnitude, whether positive or negative.}
    \label{fig: vol-tutorial}
\end{figure}

Figure \ref{fig: vol-tutorial} provides an example of the connection between price, log returns, and volatility. On the left we have a simulated set of price observations over one year, and in the center we have the associated log returns. Finally, on the right, we have the volatility path overlaid on the returns. We can see that where volatility is high we have larger returns (both positive and negative), and where volatility is low the returns tend to be small. 
Naturally if we wish to understand how the price will evolve in the future we need to also understand how volatility will evolve.

\section{Extended Methods}

\subsection{Moving Average Gaussian Processes}\label{app: magpie}

\begin{figure*}[t!]
    \centering
    \includegraphics[width=0.8\linewidth]{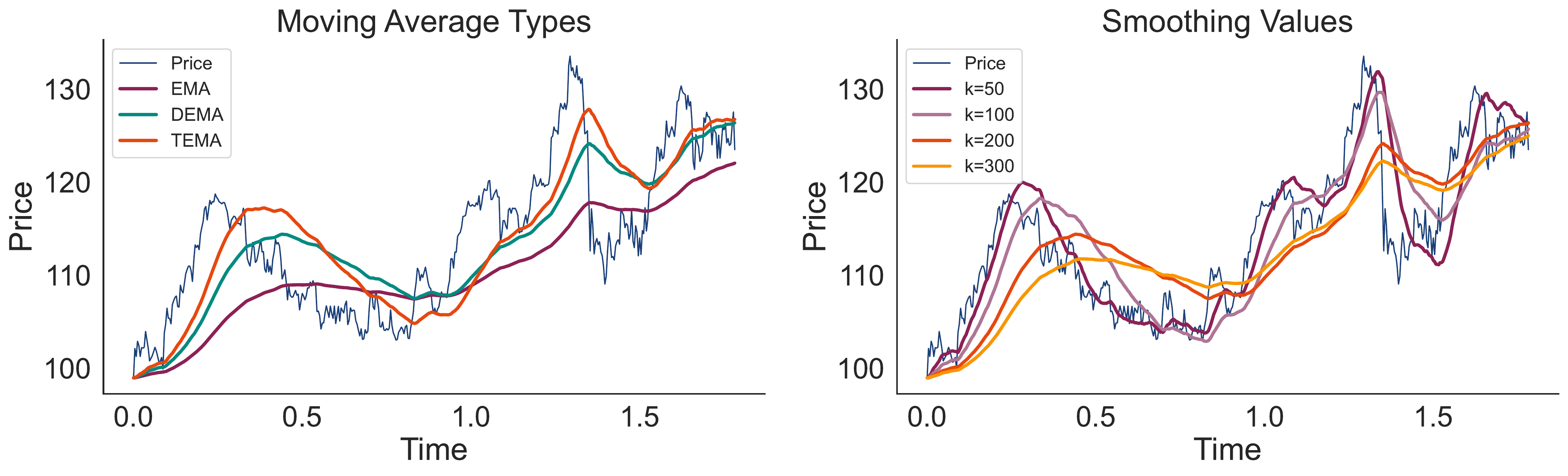}
    \caption{\textbf{Left:} a comparison of EMA, DEMA, and TEMA methods for producing moving averages for $k=200$. Note that for a fixed value of $k$ the DEMA and TEMA curves resolve a portion of the lag issue seen in the EMA curve. 
    \textbf{Right:} DEMA curves for various values of $k$. 
    Increasing $k$ averages over more historical data. }
    \label{fig: moving-average-example}
\end{figure*}

Figure \ref{fig: moving-average-example} gives an example of how various moving averages (or Magpie prior means) appear given a series of price observations for a stock. On the left, the standard EMA formulation displays a clear lag effect, that is ameliorated by using either Double or Triple moving averages (DEMA and TEMA). On the right, we see how the DEMA moving average varies for different smoothing parameters $k$; for larger values the moving average is less sensitive to fluctuations in the data, but exhibits more bias, similarly smaller values of $k$ produce moving averages that more closely match the data, but are susceptible to outliers. 

\subsection{Proofs from Derivations} \label{app: proofs}

\paragraph{Log-Volatility Kernel Function}
Recall the SDE governing movements in the log-volatility:
\begin{equation}\label{app: log-vol-sde}
dv(t) = -\frac{\sigma^2}{2}dt + \sigma dZ(t).
\end{equation}
We now derive the covariance function $\text{Cov}(v(t), v(t'))$, assuming without loss of generality, that $t < t'$. For ease of notation, and as the mean does not affect the covariance structure, let $\tilde{v}(t)$ be the same process as $v(t)$ with the mean trend removed. 

Using independence of increments of the SDE we can determine the covariance as follows:
\begin{align*}
  \text{Cov}(v(t), v(t')) &= \text{Cov} (v(t) - E[v(t)], v(t') - E[v(t')])\\
  &= \text{Cov} (\tilde{v}(t), \tilde{v}(t')) \\
  & = E[\tilde{v}(t) \tilde{v}(t')] - E[\tilde{v}(t)]E[\tilde{v}(t')] \\
  & =  E[\tilde{v}(t) \tilde{v}(t')]\\
  &= E[\tilde{v}(t) (\tilde{v}(t') - \tilde{v}(t))] + E[\tilde{v}(t')^2] \\
  &= E[\tilde{v}(t')^2] = t' \sigma^2 = \min\{t, t'\}\sigma^2.
\end{align*}

So finally we have $\text{Cov}(v(t), v(t')) = K_v(t, t') = \min\{t, t'\}\sigma^2$.

\paragraph{Log-Price Kernel Function}
\begin{equation}\label{app: log-price-sde}
ds(t) = \mu_{s}dt + V(t) dW(t)\\
\end{equation}

The covariance function of $s(t)$ can be derived using the fact that the $s(t)$ diffusion has independent increments; first assume that $t < t'$ and that $\tilde{s}(t)$ is the same process as $s(t)$ with the mean trend removed. Therefore,
\begin{align*}
  \text{Cov}(s(t), s(t')) &= \text{Cov} (s(t) - E[s(t)], s(t') - E[s(t')])\\
  &= \text{Cov} (\tilde{s}(t), \tilde{s}(t')) \\
  & = E[\tilde{s}(t) \tilde{s}(t')] - E[\tilde{s}(t)]E[\tilde{s}(t')] \\
  & =  E[\tilde{s}(t) \tilde{s}(t')]\\
  &= E[\tilde{s}(t) (\tilde{s}(t') - \tilde{s}(t))] + E[\tilde{s}(t')^2] \\
  &= E[\tilde{s}(t')^2],
\end{align*}
now since $E[\tilde{s}(t)] = 0$, $E[\tilde{s}(t)^2] = \text{Var}(\tilde{s}(t)) = \text{Var}(s(t))$ which is just the integral of the variance of the diffusion in Equation \eqref{eqn: joint-bm}, leaving us with 
\begin{align*}
    Cov(s(t), s(t')) = \int_{0}^{\min\{t,t'\}}V(t)^2 dt = K_s(t, t'; V(t)).
\end{align*}

\subsection{GPCV Training}\label{app:gpcv_training}
\paragraph{GPCV Likelihood}
As described in the main text, we model the log returns, $w(t),$ at time $t$ as independently distributed following the construction of \citet{wilson_copula_2010}.
That is, $w(t) \sim \mathcal{N}(0, \gamma^2(t)),$ where $\gamma(t)$ is the latent standard deviation.
We choose $\gamma(t) = \exp\{f(t)\},$ which is equivalent to the parameterization used in \citet{lazaro2011variational}.
The exponential parameterization has the nice property that we are also modelling the log prices in the SDE formulation described in the rest of the paper, unlike \citet{wilson_copula_2010}'s softplus transformation of the latent process.
\citet{wilson_copula_2010} also study the exponential parameterization for a few experiments.

We note that $\gamma(t)$ is a daily volatility and to convert to an annualized volatilty like in the rest of the paper, we need to rescale it by a factor of $1/\sqrt{t},$ so that $\hat \gamma(t) = \gamma(t) / \sqrt{t}.$

\paragraph{Inference Scheme}

Following \citet{hensman2013gaussian,hensman2015scalable}, we want to compute the ELBO as 
\begin{align}
    \log p(y) \geq \mathbb{E}_{q(f)}(\log p(y | f)) - \text{KL}(q(u) || p(u)),
\end{align}
where $p(y|f)$ is the GPCV volatility likelihood and $\text{KL}(q(u) || p(u))$ is the Kullback-Leibler divergence between the the variational distribution $q(u) = \mathcal{N}(m, S)$ and the prior $p(u)$.
We need to optimize $q(u)$, our free form variational distribution and estimate $\mathbb{E}_{q(f)}(\log p(y | f))$ using Bayesian quadrature as in \citet{hensman2015scalable}.

As $T$ is generally pretty small, we set the inducing points, $u,$ to be the training data points, e.g. $\{t_i\}_{i=1}^T.$
We initialize the variational mean $m$ to be the logarithm of the running standard deviation of the log returns, and the variational covariance to be $K_{uu}(K_{uu} + K_{uu}\Sigma_yK_{uu})^{-1}K_{uu}$ where $\Sigma_y$ is the negative Hessian at the initial value of $m.$

Computational and memory costs then run at about $\mathcal{O}(T^3)$ time.
In the future, we hope to use sliding windows for the inducing points, enabling mini-batching, reducing the cost to $\mathcal{O}(T_{\text{window}}^3)$ time \citep{hensman2015scalable}.
Finally, our inference scheme is simply a more flexible version of the fixed-form heteroscedastic scheme used in \citet{lazaro2011variational}, which we found to be too inflexible to fit rougher volatility paths well.

\paragraph{Multi-task Parameterization}

We follow the ICM-like model parameterization of \citet{dai2017efficient} by parameterizing $q(u) = \mathcal{N}(m, S_x \otimes S_T)$ and assume that $p(u) = \mathcal{N}(\mu(u), K_{uu} \otimes K_{TT}).$
Then we need to compute $q(f)$ which can be done for single-task models as 
$q(f) = \mathcal{N}(K_{fu}K_{uu}^{-1} m, K_{ff} + K_{fu}K_{uu}^{-1}(S - K_{uu})K_{uu}^{-1}K_{uf}).$
In the multi-task setting, this is algebraically written as:
\begin{align}
    q(f) &= \mathcal{N}((K_{fu}\otimes K_{TT}) (K_{uu}\otimes K_{TT})^{-1} m, \nonumber \\
    & (K_{ff} \otimes K_{TT}) + (K_{fu}\otimes K_{TT})(K_{uu}\otimes K_{TT})^{-1}(S_x \otimes S_t - (K_{uu}\otimes K_{TT}))(K_{uu}\otimes K_{TT})^{-1}(K_{fu}\otimes K_{TT})^\top) \nonumber \\
    &= \mathcal{N}\left((K_{fu}K_{uu}^{-1} \otimes I)m, (K_{ff} \otimes K_{TT}) + (K_{fu}K_{uu}^{-1} \otimes I)(S_x \otimes S_t - (K_{uu}\otimes K_{TT}))(K_{fu}K_{uu}^{-1} \otimes I)^\top \right) \nonumber \\
    &= \mathcal{N}\left((K_{fu}K_{uu}^{-1} \otimes I)m, (K_{ff} - K_{fu}K_{uu}^{-1}K_{uf} \otimes K_{TT}) + (K_{fu}K_{uu}^{-1} S_x K_{uu}^{-1} K_{uf} \otimes S_t) \right)
\end{align}
Note that the variational mean term is a batch matrix vector multiplication, while the variational covariance form is a sum of two Kronecker products.
Together we can sample from the posterior distribution in $\mathcal{O}(T^3 + P^3)$ time by using Kronecker identities as described in \citet{rakitsch2013all}.

In the multi-task setting, we also initialize the variational covariance to be the average initial covariance across tasks and the variational intertask covariance to be the covariance of $m$ across tasks.
The intertask covariance is a $P \times P$ matrix parameterized as rank one plus diagonal; we regularize it with a LKJ prior with $\eta = 5.0$ \citep{lewandowski2009generating}.

Additionally, we exploit Kronecker identities to efficiently compute the KL divergence in the variational distribution so that training stays at $\mathcal{O}(T^3 + P^3)$ time by broadly following the approach of \citet{dai2017efficient}.

\subsection{Model Training}

All models were trained in GPyTorch \citep{gardner2018gpytorch} and PyTorch \citep{paszke2019pytorch} on either a single $24$GB GPU or a single $12$GB GPU; the multi-task wind experiment used a $48$GB Titan RTX GPU.
Training time was negligible, with models typically taking less than $1$ minute to train.
For training, we use $500$ steps of Adam with learning rate $0.1$ and optimize through the log marginal likelihood.

\paragraph{Multitask GPs}
We use the ICM model of \citet{bonilla2007multi}.
Like in the GPCV setting, we use a rank one plus diagonal intertask covariance, regularized with a LKJ prior \citep{lewandowski2009generating}.
By structure exploitation, these models cost $\mathcal{O}(P^3 + T^3)$ for fitting and $\mathcal{O}(P^3 + T^3)$ for posterior sampling when using Matheron's rule \citep{maddox2021bayesian}.

\paragraph{Data Space GPs}
We use a standard Gaussian likelihood for these responses on the log transformed data and optimize both the scale of the volatility as well as the noise term, initializing the noise to be $10^{-4}.$
As these models reduce to a standard exact GP conditional on volatility, computational and memory costs then run at $\mathcal{O}(T^3)$ time.

\section{Experimental Details}\label{app: exps}

\subsection{Details from Section \ref{sec:stocks}}\label{app: forecasting}

We source daily closing prices for stocks in the Nasdaq $100$ for $2$ years prior to January $2022$. Volt models are trained according to the outline in Section \ref{sec: inference}, and standard GPs are implemented and trained via GPyTorch and BoTorch \citep{gardner2018gpytorch, balandat2019botorch}. The LSTM model is implemented with $2$ hidden layers each with $128$ units and takes the form
\begin{align*}
f(s_t, s_{t-1}, s_{t-2}, s_{t-3}, s_{t-4}) = \{\hat{\mu}_{t+1} \hat{\sigma}_{t+1}\}
\end{align*}
where $\hat{\mu}_{t+1}$ is the predicted mean at time $t+1$, and $\hat{\sigma}_{t+1}$ is the predicted standard deviation at time $t+1$.

For each stock in our universe we select $25$ cutoff times at which we generate forecasts, using the preceding $400$ observations as training data.
At each cutoff time we forecast the log closing price $100$ days into the future, and compute the calibration and negative log likelihood of the forecasts $75$ to $100$ days out. We specifically focus on longer horizon forecasts, as it is generally a harder task for which out of the box methods are ill-suited.

\begin{table}[h!]
\begin{small}
    \centering
    \begin{tabular}{c|c c c }
         & Stock Prices & Wind Speeds & FX \\
         \hline
         Volt + Magpie & $5.88 \pm 0.02$ & $4.28 \pm 0.16$ & $-1.69 \pm 0.02$\\
         
         Volt + Con. &  $\mathbf{4.69 \pm 0.03}$ & $\mathbf{3.38 \pm 0.05}$ & $-1.60 \pm 0.02$\\
         
         Mat\'ern + Magpie & $9.80 \pm 0.27$ & $12.13 \pm 0.81$& $4.23 \pm 0.30$\\
         
         Mat\'ern + Con. & $7.74 \pm 0.21$ & $18.03 \pm 1.90$ & $-0.36 \pm 0.04$\\
         
         SM + Magpie & $147.84 \pm 1.84$ & $110.07\pm 7.81$ & $562.67 \pm 15.71$\\
         
         SM + Con. & $80.43 \pm 0.57$ & $70.14\pm 5.03$ & $356.55 \pm 11.98$\\
         
         LSTM & $49.95 \pm 0.59$ & $45.13\pm 1.82$ & $10.66 \pm 0.44$\\
         
         Volt-VHGP + Con. & $4.76 \pm 3.05$ & $5.75 \pm 0.44$ & $-1.58 \pm 0.03$\\
         
         Volt-VHGP + Magpie& $6.97 \pm 1.24$ & $5.91 \pm 0.34$ & $-1.66 \pm 0.02$\\
         
         GPCV & $5.45 \pm 1.51$ & $4.89 \pm 0.04$ & $\mathbf{-1.79 \pm 0.02}$\\
         \bottomrule
    \end{tabular}
    \caption{Negative log likelihoods (NLLs) per test point for the methods compared on both the stock forecasting and wind speed tasks, averaged of tens of thousands of forecasts. While there is a slight improvement in NLL from using a constant mean, the inclusion of Magpie is central to achieving high calibration.}
    \label{tab:nlls-ext}
\end{small}
\end{table}

\subsection{Details from Section \ref{sec: wind}}\label{app: wind}

We source data from \citet{diamond2013us} for the $2021$ calendar year. Wind measurements are taken at $15$ minute intervals for all $154$ stations in the observation network. In order to treat the observed wind speed as log-normally distributed we add $1$ to each observation (to shift the $0$ m/s observations to a value of $1$), and then model the log of the resulting time series.

\begin{figure}
    \centering
    \includegraphics[width=0.5\linewidth]{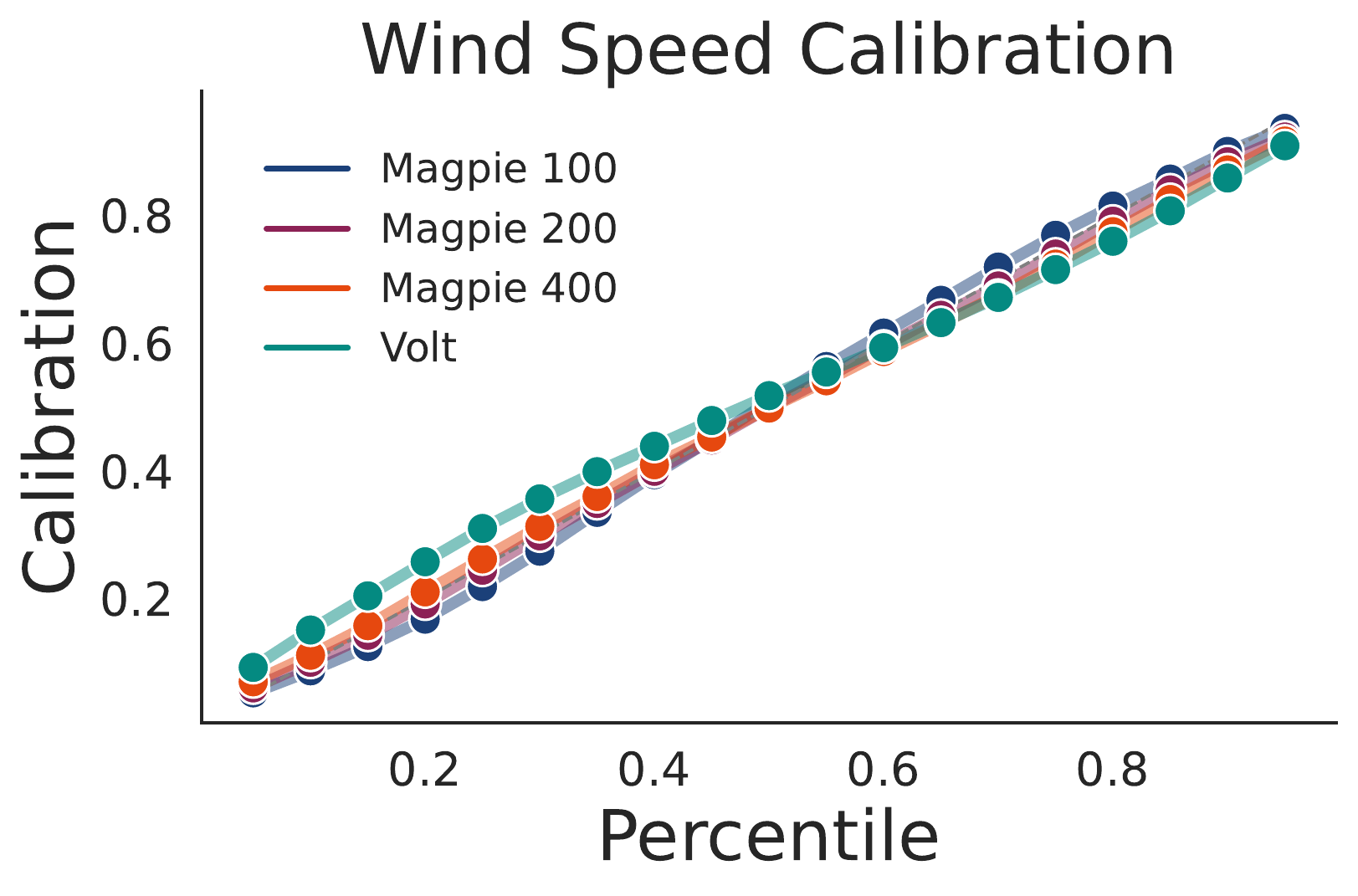}
    \caption{A comparison of Volt with a constant mean, and Volt with various Magpie means in terms of calibration in wind forecasts. While Volt with a constant mean is well calibrated, it is aided by the inclusion of a Magpie mean with large smoothing parameter.}
    \label{fig:wind-ema}
\end{figure}

Figure \ref{fig:wind-ema} compares the performance of Volt alone and Volt with Magpie mean functions with various smoothing parameters. Magpie means aid in calibration, although the effect is less pronounced as we see with stock forecasting in Figure \ref{fig:stock-calib}.

A key distinction between wind speed forecasting and stock price forecasting is that wind speeds tend to revert to a consistent level, whereas stock prices may increase by thousands then stabilize at a new level. For this reason we explore the use of mean reversion in our rollout forecasts. To add mean reversion to the rollouts we simply adjust the posterior mean of the GP towards the mean of the training data by a factor of $\theta$. 
That is, rather than sampling from the GP posterior $s_{t^*} \sim \mathcal{N}(\mu^*_{f | \mathcal{D}} | \Sigma^*_{f | \mathcal{D}})$ we sample from $s_{t^*} \sim \mathcal{N}(\mu^*_{f | \mathcal{D}} - \theta (\mu^*_{f | \mathcal{D}} - \frac{1}{N}\sum_i s_i) | \Sigma^*_{f | \mathcal{D}})$. 

\begin{figure}
    \centering
    \includegraphics[width=0.5\linewidth]{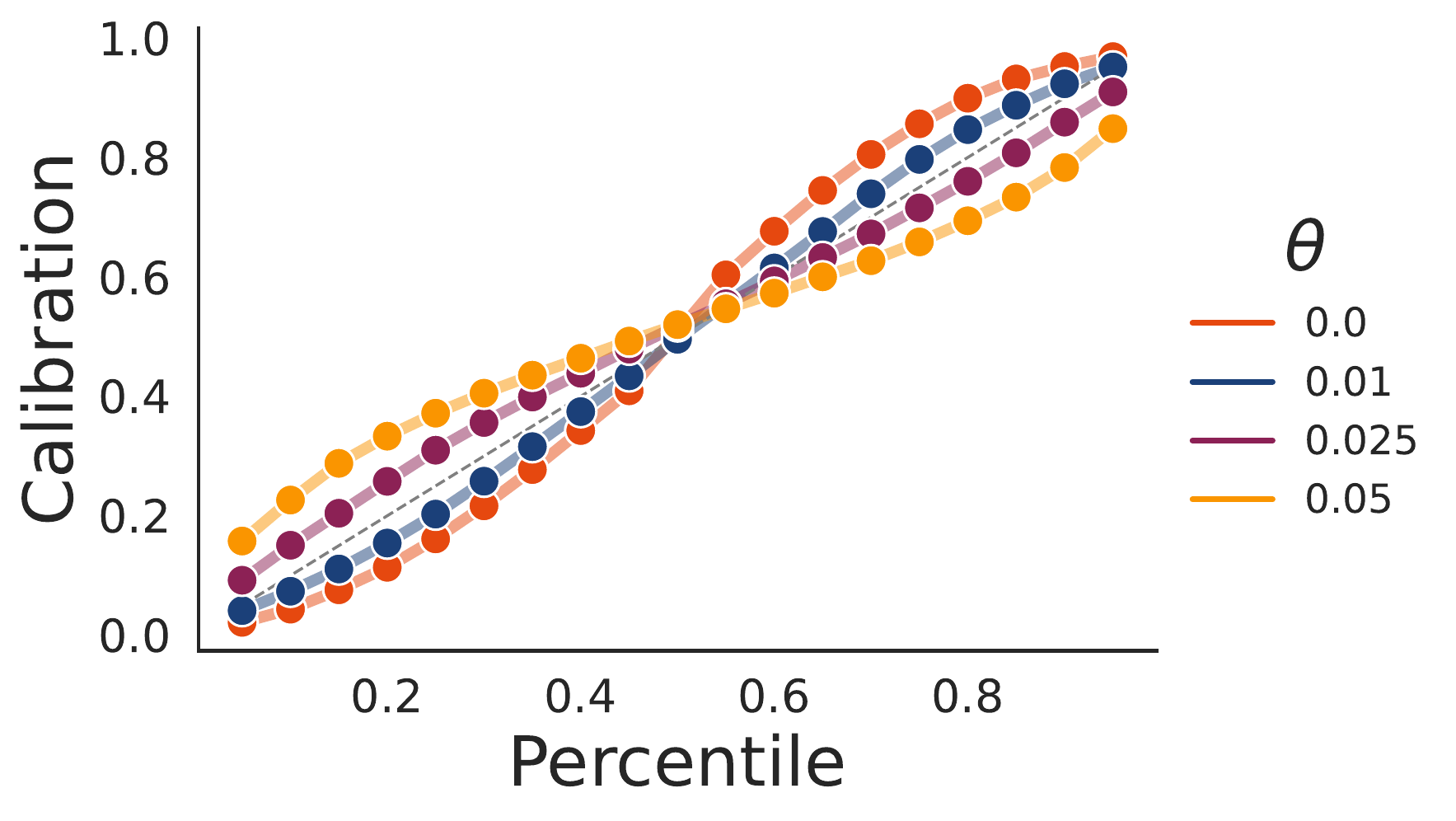}
    \caption{Calibration of different mean reversion $\theta$ values across stock prices.}
    \label{fig:theta}
\end{figure}

In this mean reversion setting, $\theta$ controls the speed at which rollouts tend to revert towards the mean. At $\theta = 0$ we are in the standard GP prediction case, at $\theta=1$, we only ever sample from a distribution centered around the mean of the training observations. Figure \ref{fig:theta} provides a comparison of the calibration under differing levels of mean reversion for Volt. The standard Volt rollouts are in general well calibrated for this problem, but we see that just a small amount of mean reversion can increase the overall calibration notably.  

\subsection{Details from Section \ref{sec: multitask}} \label{app:mt_volt_experiments}

\begin{figure*}[h!]
    \centering
    \includegraphics[width=\textwidth]{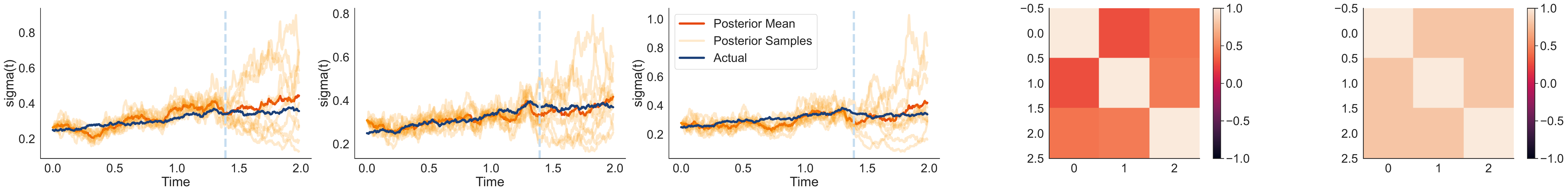}
    \caption{\textbf{Left three panels:} Predicted correlated volatility models. \textbf{Fourth panel:} Estimated correlation matrix between volatility models. 
    \textbf{Fifth panel: } True volatility correlation. Multitask GPCVs are able to both predict volatility while also estimating the true correlation between volatility.}
    \label{fig:mt_gpcv}
\end{figure*}

In Figure \ref{fig:mt_gpcv}, we construct a multi-task SABR volatility model with correlations given by the farthest right panel and volatility processes given as the blue lines in the left three panels.
We then use our multi-task GPCV model to estimate and predict the true volatilities for each task in the left three panels, while also estimating the true relationships between each volatility. 
The estimated relationships are shown in the fourth panel from left, which is pleasingly similar to the true correlation shown at far right.

\begin{figure*}[!h]
\includegraphics[width=\linewidth]{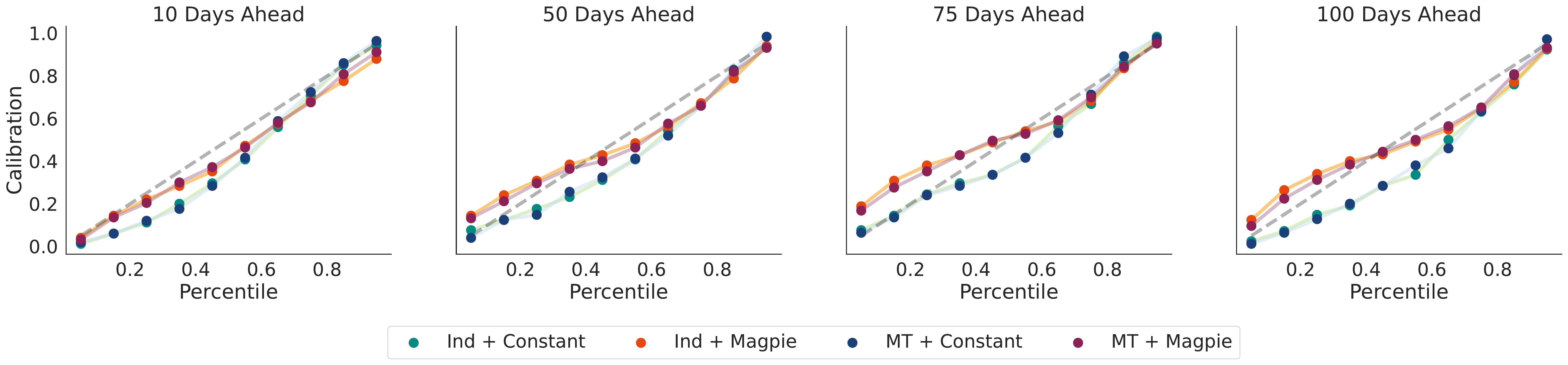}
\caption{Calibration of multi-task Volt and independent models across time step lookaheads for $5$ different SPDRS.}
\label{fig:mtspdrs_calibration}
\end{figure*}

\begin{figure*}
\centering
\begin{subfigure}{0.38\textwidth}
    \includegraphics[width=\linewidth]{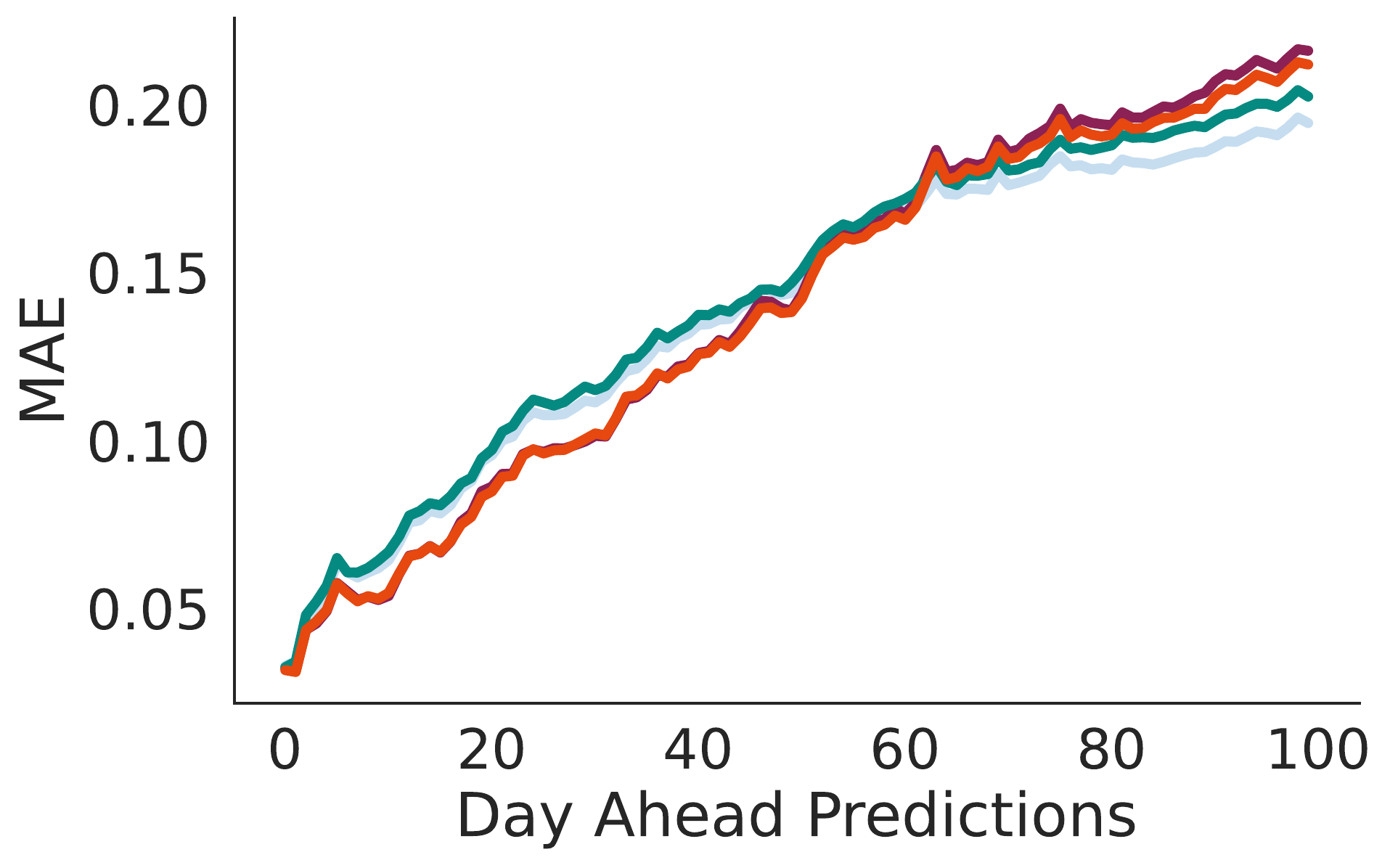}
    \caption{MAE}
    \label{fig:mtspdrs_mae}
\end{subfigure}
\begin{subfigure}{0.55\textwidth}
    \includegraphics[width=\linewidth]{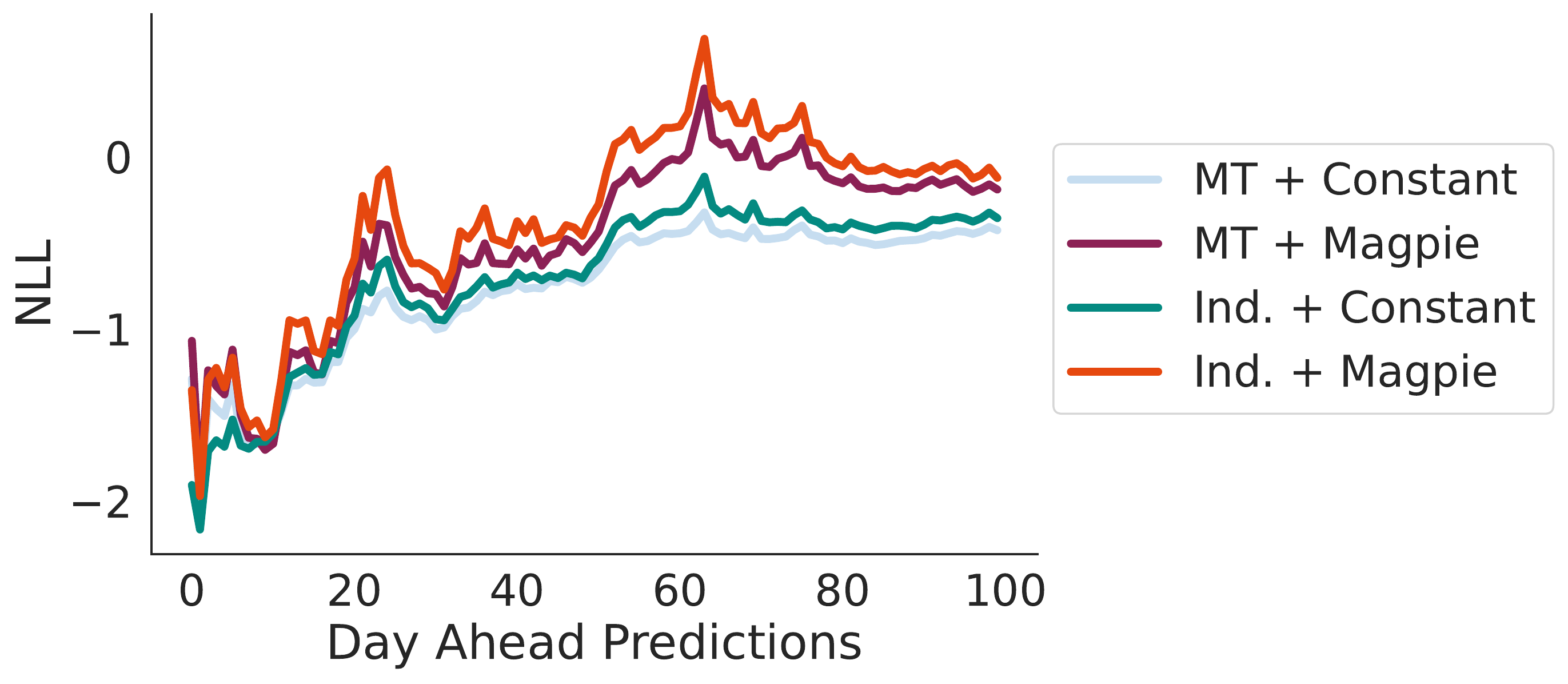}
    \caption{NLL}
    \label{fig:mtspdrs_nll}
\end{subfigure}
\caption{\textbf{Left panel:} Mean absolute error of rollouts. \textbf{Right panel: } Negative log likelihood of rollouts.}
\end{figure*}

For Figure \ref{fig:mt_volt_experiments}left and Figures \ref{fig:mtspdrs_calibration}, \ref{fig:mtspdrs_mae}, \ref{fig:mtspdrs_nll}, we fit stocks comprising of five different exchange traded fund SPDRs\footnote{\url{https://en.wikipedia.org/wiki/SPDR}} collected over $5$ years of daily data from $09/2016$ to $09/2021$. These SPDRS are XLE, XLF, XLK, XLRE, XLY; each had six stocks in it except for XLF which had $30$.
We fit on $300$ days and evaluated $100$ days into the future, with $5$ rolling testing sets for each prediction.

For Figure \ref{fig:mt_volt_experiments} center, we used the same training data except used only five stocks from the XLE SPDR and five from the XLF SPDR.

\begin{figure*}
\centering
\includegraphics[width=0.75\linewidth]{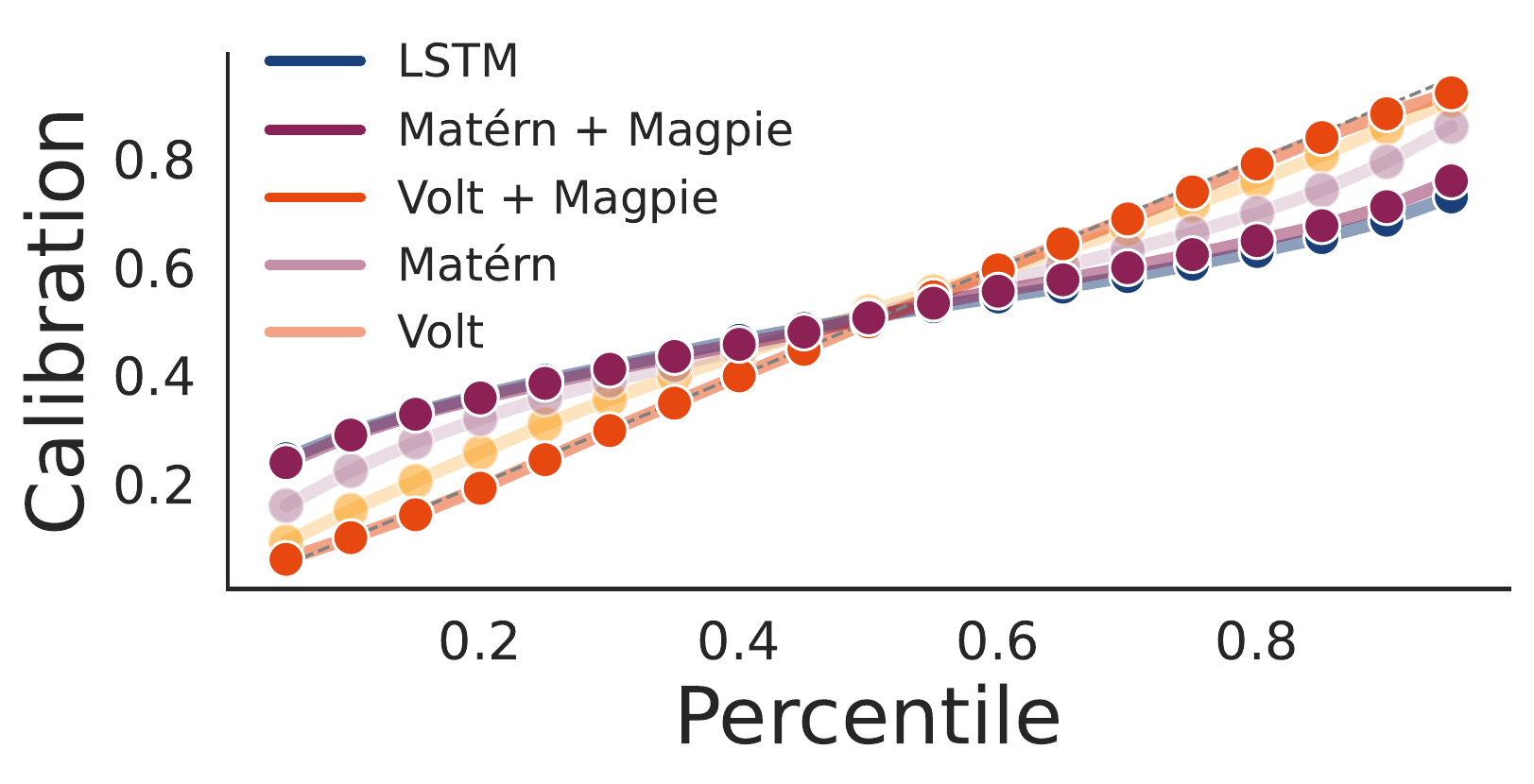}
\caption{Calibration of multi-task Volt and independent models across time step lookaheads for the wind forecasting datasets.}
\label{fig:mtwind_calibration}
\end{figure*}

\begin{figure}[ht]
    \centering
    \includegraphics[width=0.5\linewidth]{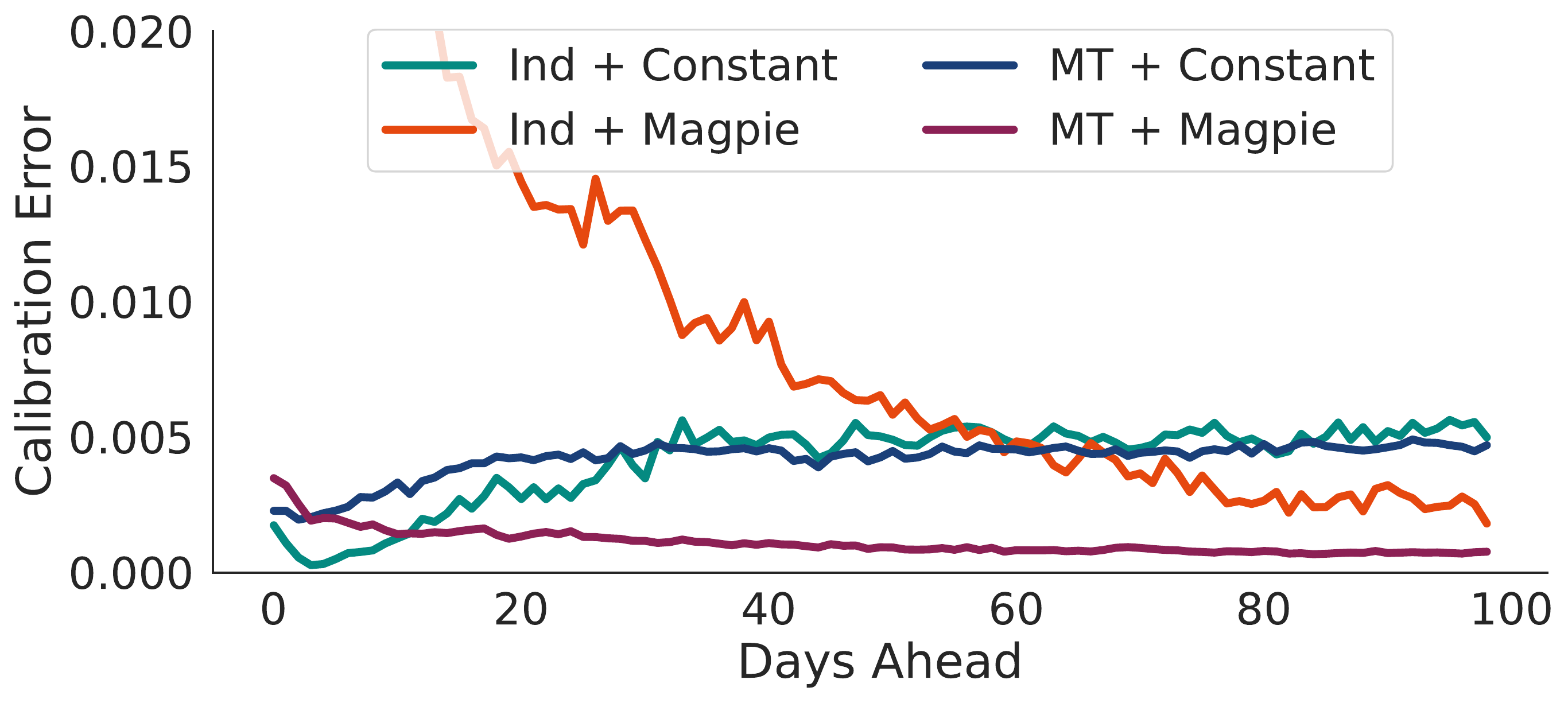}
    \caption{Calibration error of the models across different time step lookaheads for the wind forecasting task.}
    \label{fig:mtwind_ece}
\end{figure}

For Figure \ref{fig:mt_volt_experiments} right and Figures \ref{fig:mtwind_calibration}, \ref{fig:mtwind_ece} we fit about $100$ different wind stations (depending on amount of missing data) at $5$ minute intervals across $2021$ with $25$ indepdendent rolling splits.
We fit on $252$ increments and tested on $100$ increments.
Here, on the multi-task ones, we used a larger RTX 8000 GPU. 



\end{document}